\documentclass[sigconf,nonacm]{acmart}

\AtBeginDocument{%
  \providecommand\BibTeX{{%
    \normalfont B\kern-0.5em{\scshape i\kern-0.25em b}\kern-0.8em\TeX}}}

\setcopyright{acmlicensed}
\copyrightyear{2024}
\acmYear{2024}
\acmDOI{XXXXXXX.XXXXXXX}

\acmConference[KDD 2025]{KDD 2025}{August 3--7,
  2025}{Toronto, ON, Canada}
\usepackage{dsfont}
\usepackage{comment}
\usepackage{multirow}
\usepackage{luca}

\newcommand{\batch}{{X}}
\newcommand{\items}{{\mathcal{I}}}

\newcommand{\pipeline}{\textsc{DriftInspector~}}
\newcommand{\pipelinev}{\textsc{DriftInspector}}

\newcommand{\colorA}[1]{ {\color{magenta} #1} }
\newcommand{\colorB}[1]{ {\color{violet} #1} }
\newcommand{\colorC}[1]{ {\color{orange} #1} }
\newcommand{\colorD}[1]{ {\color{teal} #1} }

\newcommand{\minsup}{s}

\def\sat{\models}			% |=
\def\set#1{\{ #1 \}}			% { #1 }

\newcommand{\divexp}{\textsc{DivExplorer}}

\newcommand{\rev}[1]{{\color{black} #1}}
\newcommand{\new}[1]{{\color{black} #1 }}

\title{Detecting Interpretable Subgroup Drifts}

\author{Flavio Giobergia}
\authornote{Corresponding author -- flavio.giobergia@polito.it}
\affiliation{%
  \institution{Politecnico di Torino}
  \city{Turin}
  \country{Italy}
}
\author{Eliana Pastor}
\affiliation{%
  \institution{Politecnico di Torino}
  \city{Turin}
  \country{Italy}
}
\author{Luca de Alfaro}
\affiliation{%
  \institution{University of California, Santa Cruz}
  \city{Santa Cruz}
  \country{USA}
}
\author{Elena Baralis}
\affiliation{%
  \institution{Politecnico di Torino}
  \city{Turin}
  \country{Italy}
}

\renewcommand{\shortauthors}{Author 1 al.}
\newcommand{\githubrepo}{ \url{https://anonymous.4open.science/r/drift-inspector/}}

\begin{document}

\begin{abstract}
The ability to detect and adapt to changes in data distributions is crucial to maintain the accuracy and reliability of machine learning models. 
Detection is generally approached by observing the drift of model performance from a global point of view. 
However, drifts occurring in (fine-grained) data subgroups may go unnoticed when monitoring global drift.
We take a different perspective, and introduce methods for observing drift at the finer granularity of subgroups. 
Relevant data subgroups are identified during training and monitored efficiently throughout the model's life. 
Performance drifts in any subgroup are detected, quantified and characterized so as to provide an interpretable summary of the model behavior over time. 
Experimental results confirm that our subgroup-level drift analysis identifies drifts that do not show at the (coarser) global dataset level.
The proposed approach provides a valuable tool for monitoring model performance in dynamic real-world applications, offering insights into the evolving nature of data and ultimately contributing to more robust and adaptive models.
\end{abstract}

% This avoid lines that are too long. 
\sloppy



\maketitle

\section{Introduction}
\label{sec:intro}

\begin{figure}
    \centering
    \includegraphics[width=1\linewidth]{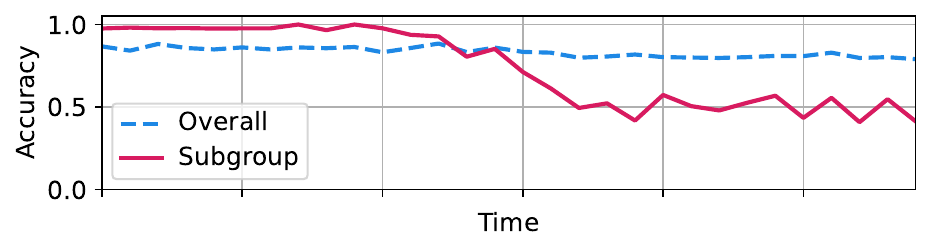}
    \caption{
    \rev{Adult dataset. Drift event affecting the subgroup \textit{women under the age of 36, employed in the private sector} (approx. 10\% of the overall population). The overall accuracy is only marginally affected, whereas the drop in performance for the drifting subgroup is more prominent. 
    % Global drift detection methods struggle to detect this type of drift. 
    }
    % ; overall accuracy and for a drifting subgroup over time. We discuss this experiment on the CelebA~\cite{liu2015faceattributes} dataset in Section~\ref{sec:experimentalresults}.
    }
     \Description{Example of a drifting event}
    \label{fig:example-running}
\end{figure}

%Machine learning models are increasingly adopted in a wide range of applications. 
Given the rapid evolution of data, maintaining machine learning model accuracy and reliability is critical. 
% In this dynamic landscape, d
Detecting and adapting to shifts in performance ensure that models remain effective over time.
Traditionally,  the approaches for detecting drifts in model performance take a global perspective, focusing on overall performance. However, changes in the data distribution or in the relationship between input and target variables may not be widespread across all data, but may affect specific subpopulations. Hence, traditional methods may overlook finer-grained drifts occurring in specific data subgroups.

\rev{Consider as an example the Adult dataset~\cite{misc_adult_2}. 
To predict the income bracket of a person based on their demographic information, a classification model can be trained on the available data.
This data represents a snapshot of the socioeconomic conditions at training time. 
However, these conditions may be subject to changes.
Consider, for instance, the (relatively small) subgroup including \textit{women under the age of 36, employed in the private sector}. 
The subgroup income bracket is initially predicted with near-perfect accuracy by the classification model. 
Over time, the subgroup exhibits a new behavior: the salary for that segment increases, as the gender pay gap is reduced.
This behavior is depicted in Figure~\ref{fig:example-running}. While the overall model performance (dashed blue line) is only marginally affected by the drift, the subgroup performance (red line) changes significantly over time.
Drift detection methods that only focus on global performance may fail to detect this drift.
}

% We visually represent this phenomenon in Figure~\ref{fig:example-running}. 
% While the overall accuracy changes only slightly, the accuracy within a specific subgroup may undergo a significant drop.
\rev{
Identifying subgroup drifts enables assessing whether the model remains reliable and accurate across subpopulations. Thus, it also allows detecting model unfairness due to drift, as changes in performance may disproportionately affect different demographic groups or other protected subpopulations. Finally, it allows targeted interventions to prevent any subgroup from experiencing disservice or degraded performance over time.}

We propose \pipelinev, a methodology that identifies relevant subgroups during model training, on the basis of the interpretable features of the data. It aims to (i) detect drift in data subgroups, (ii) describe the subgroups affected by drift in an interpretable manner, and (iii) quantify the drift affecting them. 
%Consider, for example, a credit scoring model adopted to support decisions for loan applications. 
%Over time, it may suffer from a drift in performance. 
%Our techniques allow us to detect the subgroups affected by the drift, describe them, and quantify the drift affecting them. 
% Rather than only detecting that a drift occurred, be it overall or for a subpopulation of the data, we want to describe the characteristics of the instances for which it occurred. 
%For example, we might detect a significant drop in performance for young adults from a specific region. 
%With this information, experts could investigate if changes in the economic conditions or spending habits of this group affect performance.
We develop algorithms that allow the efficient monitoring of subgroup performance throughout model lifetime, enabling the timely detection and quantification of subgroup-level performance drifts. 

We characterize each data instance through interpretable features, possibly extracted from binary data.
For instance, in image data, the interpretable features can include metadata such as the image source, location, creation date, features derived from the caption, or high-level descriptors like whether the photo was taken indoors or outdoors. 
Similarly, for tabular data, we can use as features either the attribute values themselves, or coarser %or hierarchical 
representations to enhance interpretability (e.g., aggregating precise income values into ranges). 
% from age equal to 20 to `young').
%

\pipeline produces an interpretable summary of the subgroup-level drifts that can be used to guide model improvements and interventions. 
Considering the example in Figure~\ref{fig:example-running}, our method detects a significant drop in performance for
women under the age of 36 employed in the private sector. Experts could then decide the most appropriate action (e.g., retrain the model). With a global drift detection method, this drop would be hidden by the stable performance at the global level, thus delaying any action to address the performance issue.
Our experiments demonstrate the effectiveness of our approach in identifying performance drifts that occur in specific subgroups.

\rev{Our approach is performance-based~\cite{bayram2022concept} and detects drift by considering changes in performance over time, as new labels become available.
% This scenario of progressive label availability is common in multiple applications, such as spam detection applications, where users label mail as spam or not spam; loan approval, where we have the loan decision, traffic management, where we can monitor real-time traffic conditions; online advertisement, where we can have near-real-time feedback on user engagement; and social media content moderation, where users and moderators can flag content as appropriate or not.
This scenario of progressive label availability is common in multiple applications such as spam detection, loan approval, traffic management, online advertising, and social media content moderation. For example, %in traffic management, we can monitor real-time traffic conditions; 
in online advertising, near-real-time feedback on user engagement (e.g., clicks on an ad or purchases following ad views) is typically available. %Unlike traditional methods that detect global drift, our approach focuses on identifying performance drifts in subgroups.
}

Our main contributions are as follows. 
\rev{
\begin{itemize}
    \item \textbf{Subgroup-level drift definition.} We introduce the notion of \emph{subgroup-level drift}, as distinct from the global drift that affects a model in its entirety. 
    \item \textbf{Fine-grained drift detection and monitoring.} We present a novel approach to model performance monitoring over subgroups, based on sparse matrix multiplication and efficient tests of statistical significance.
    Our method is able to identify drifts at the subgroup level that would not have been visible via global-level drift monitoring only.
    \item \textbf{Effective drift exploration.} We characterize subgroups via interpretable representations provided by itemset mining techniques. Interpretable subgroups allow the investigation of the factors contributing to performance drifts.   
    \end{itemize}
}
The rest of this paper is organized as follows.
Section~\ref{sec:rw} reviews the related works.
Section~\ref{sec:backgnotation} introduces the background and the problem definition.
Section~\ref{sec:drift} describes our subgroup-based drift approach. 
Section~\ref{sec:exp} presents the experimental results.
Finally, in Section~\ref{sec:conclusion},  we draw conclusions and outline future directions.

\section{Related work}
\label{sec:rw}
We analyze related works from two different perspectives: (i) drift detection, and (ii) subgroup analysis.

%\subsection{Drift detection}
%\smallskip
\paragraph{Drift detection.}
The main approaches for detecting drift are data distribution-based and performance-based. 
The former evaluate the similarity of the data distributions over time (e.g., \cite{kifer2004detecting, gozuaccik2021concept, dasu2006information}), while the latter focus on tracing differences in performance to detect changes~\cite{bayram2022concept}.
Our work aligns with performance-based approaches.

%, which is the predominant line in the field.

%Several approaches monitor performance and adopt statistical tests to detect statistically significant performance degradations.
Several performance-based approaches, such as DDM~\cite{gama2004learning} and its modification %EDDM~\cite{baena2006early}, %RDDM~\cite{barros2017rddm} and
HDDM~\cite{frias2014online}, 
are based on the statistical process control criterion and adopt statistical tests to identify significant performance degradations, while others directly perform statistical tests such as the Page-Hinkley~\cite{pagetest1954, mouss2004test}, Chi-squared test~\cite{Pearson1900} and/or Fisher’s Exact Test (FET)~\cite{fisherexact1922} comparing to a reference distribution.
%The widely adopted Drift Detection Method (DDM)~\cite{gama2004learning}  analyzes the error rate of a classifier, treating the error as a Bernoulli random variable with Binomial distribution.
Other approaches adopt a window technique, monitoring performance on different time windows and comparing them with a reference window~\cite{bifet2007learning, pesaranghader2016fast, nishida2017detecting, de2018wilcoxon, raab2020reactive} (e.g., ADWIN~\cite{bifet2007learning}, KSWIN~\cite{raab2020reactive}). 
%The method ADaptive WINdowing (ADWIN)~\cite{bifet2007learning} is one of the most widely known in this category. 
%This sliding window approach monitors the average of a statistical measure (e.g., the error rate) in two sub-windows within a window and detects a drift if the difference is above a threshold.  It also dynamically adapts the window size, enlarging it when there is no change and shrinking it when a change is detected.  
We also employ a sliding window approach by comparing the current window under observation with a reference window and applying a statistical test. However, we focus on performance \textit{at the subgroup level} rather than overall. This approach allows us to identify statistically significant drifts for data subgroups that may be undetected when focusing solely on the overall dataset.

% OLD
%Subgroup drift is currently an underexplored area in the drift detection literature.
%Mandoline~\cite{chen2021mandoline}, differently from us, is a distribution-based approach that leverages user-specified data slices identified by practitioners as potentially affected by a distribution change. 
%Differently from~\cite{chen2021mandoline}, we automatically detect the relevant subgroups (i.e., data slices), instead of relying on user identification. Moreover, our focus is on detecting drift in subgroup performance, rather than using subgroups to estimate model performance.

\rev{Detecting local drift in regions of the feature space is an underexplored area in drift detection~\cite{lu2018learning}. 
Current methods exploring local drifts mostly focus on unsupervised methods to detect changes in input data and distribution-based approaches. In contrast, we propose a performance-based approach to identify changes in the relationship between input data and a target variable.
Mandoline~\cite{chen2021mandoline} relies on user-specified data slices identified by practitioners. Instead, we automatically detect relevant subgroups (i.e., data slices) and focus on drifts in subgroup performance rather than using subgroups to estimate model performance.
\cite{liu2020concept} proposes a k-means space partitioning for histogram-based distribution change detection. Clustering is also adopted in~\cite{zhao2023unsupervised} to partition the data into clusters, followed by learning a discriminator for each cluster to detect drift. Clusters group the data into non-interpretable and non-overlapping instances. Our data subgroups may overlap and are interpretable (e.g., loan applications of young adults and young female adults), allowing for their inspection and analysis.
LDD-DSDA~\cite{liu2017regional} monitors differences in the regional density of the data to detect local drifts. Rather than data density, we focus on changes in performance.
\cite{heyden2024adaptive} learns an encoder-decoder model and monitors its loss using an adaptive window size and statistical testing, identifying subspaces by attribute subsets. Instead, we slice the data via attribute-value pairs and do not require additional training.}

%\subsection{Subgroup analysis}
%\smallskip
\paragraph{Subgroup analysis.} Multiple subgroup analysis techniques have been proposed to evaluate subgroup performance. % for model evaluation and debugging. 
Several approaches
%especially in fairness assessment, 
rely on the a priori knowledge of the subgroups to inspect~\cite{kahng2016visual, morina2019auditing, baylor2017tfx, saleiro2018aequitas}. 
The automatic identification of subgroups has been the focus of recent research~\cite{cabrera2019fairvis, chung2019slice, pastor2021looking, sagadeeva2021sliceline, pastorahierarchical2023}.  
We follow the approach of \cite{chung2019slice, pastor2021looking, sagadeeva2021sliceline}, slicing the dataset and representing subgroups as itemsets, i.e., conjunctions of attribute-value pairs, thus yielding interpretable subgroups.
While all these previous works characterize system performance at the level of subgroups, we focus here on monitoring how the model performance over subgroups changes in time. 

\section{Problem definition}
\label{sec:backgnotation}
We consider a model $f: \mathcal{X} \rightarrow \mathcal{Y}$ which maps elements $x \in \mathcal{X}$ of the input domain $\mathcal{X}$ to a prediction $\tilde{y} = f(x)$. The problem addressed by $f$ may encompass generic tasks such as classification or regression. %, or object detection.

We consider performance metrics that can be measured on collections $\batch = x_1, \ldots, x_N$ of input instances for $f$, and that can be expressed as the fraction of positive occurrences of an event of choice. 
Precisely, we assume that we have two indicator functions $\alpha, \beta: \mathcal{X} \mapsto \set{0, 1}$, such that $\alpha(x) = 1$ if $x$ is a positive instance, and $\beta(x) = 1$ if $x$ is a negative instance. 
Given a collection $\batch = x_1, \ldots, x_N$ of inputs, we define a performance metric $h$ via: 
\begin{equation} \label{eq-h}
    h(\batch) = \frac{\sum_{x \in \batch} \alpha(x)}{\sum_{x \in \batch} \alpha(x) + \beta(x)} \eqpun . 
\end{equation}
By taking $\alpha(x) = 1$ for correctly classified samples and $\beta(x) = 1$ for incorrectly classified ones in (\ref{eq-h}), $h$ expresses accuracy. 
By taking $\alpha(x) = 1$ for false positives and $\beta(x) = 1$ for true negatives, $h$ expresses the false-positive rate. 
We are interested in detecting whether the value of the performance metric $h$ drifts, in statistically significant ways, when measured over data subgroups. 
\rev{We assume that the ground truth labels become gradually available over time, allowing us to track the performance metric during model lifetime.}

To define subgroups, we assume that with each instance $x_i$ is associated an interpretable description, or \textit{metadata}, $D_i$. 
For images, the metadata can consist in the image source, creation date, dominant color, camera name, or location. 
For tabular data, the instance itself can be adopted as its own description. 
We assume that each piece of metadata in $D_i$ has the form of an \emph{attribute=value} pair. 
Further, we assume that the metadata can be extracted from the instance via a function $e$ of choice (i.e., $D_i = e(x_i)$), so that while extracting the metadata, we can ensure it is discretized and aggregated in ways that facilitate its interpretation. 
For example, while $x_i$ can contain a timestamp with 1-second resolution, the metadata $e(x_i)$ may contain the month on which image $x_i$ was taken, or the day of the week. 
An example of metadata for an image is \{\textit{age=50, gender=female, smiling=true, hair=blonde}\}.

Subgroups are defined on the basis of the instance metadata. 
Borrowing terminology from frequent pattern mining \cite{tan2018introduction}, each \emph{attribute=value} pair constitutes an \emph{item}; we denote by $\items$ the set of all such items. 
Subgroups are defined via \emph{itemsets}, or sets of items. 
For instance, an itemset for the above image dataset can be \{\textit{age=30, gender=female}\}.
An instance $x_i \in \mathcal{X}$ is \textit{covered} by (or satisfies) an itemset $S$ if $S \subseteq e(x_i) = D_i$, which we abbreviate as $S \subseteq x_i$.
Given a collection $\batch$, we denote with $\batch(S)$ the subcollection of $\batch$ that satisfies the itemset $S$, that is, $\batch(S) = (x \mid x \in \batch, S \subseteq x)$. 
Further, we define the support of $S$ in $\batch$, denoted $\sup_\batch(S)$, as the fraction of instances satisfying $S$, i.e. $\sup_\batch(S) = |\batch(S)| / |\batch|$.

\section{Method}
\label{sec:drift}
Our goal is to monitor the subgroup evolution and detect any drifts in the chosen performance metric over time.  
We do so by identifying a reference window of data with the expected behavior. 
We then consider subsequent windows of data (e.g., batches of data collected after deployment), and we detect statistically significant drifts in performance separately for each subgroup. 

\begin{figure*}[ht]
    \centering
    \includegraphics[width=\linewidth]{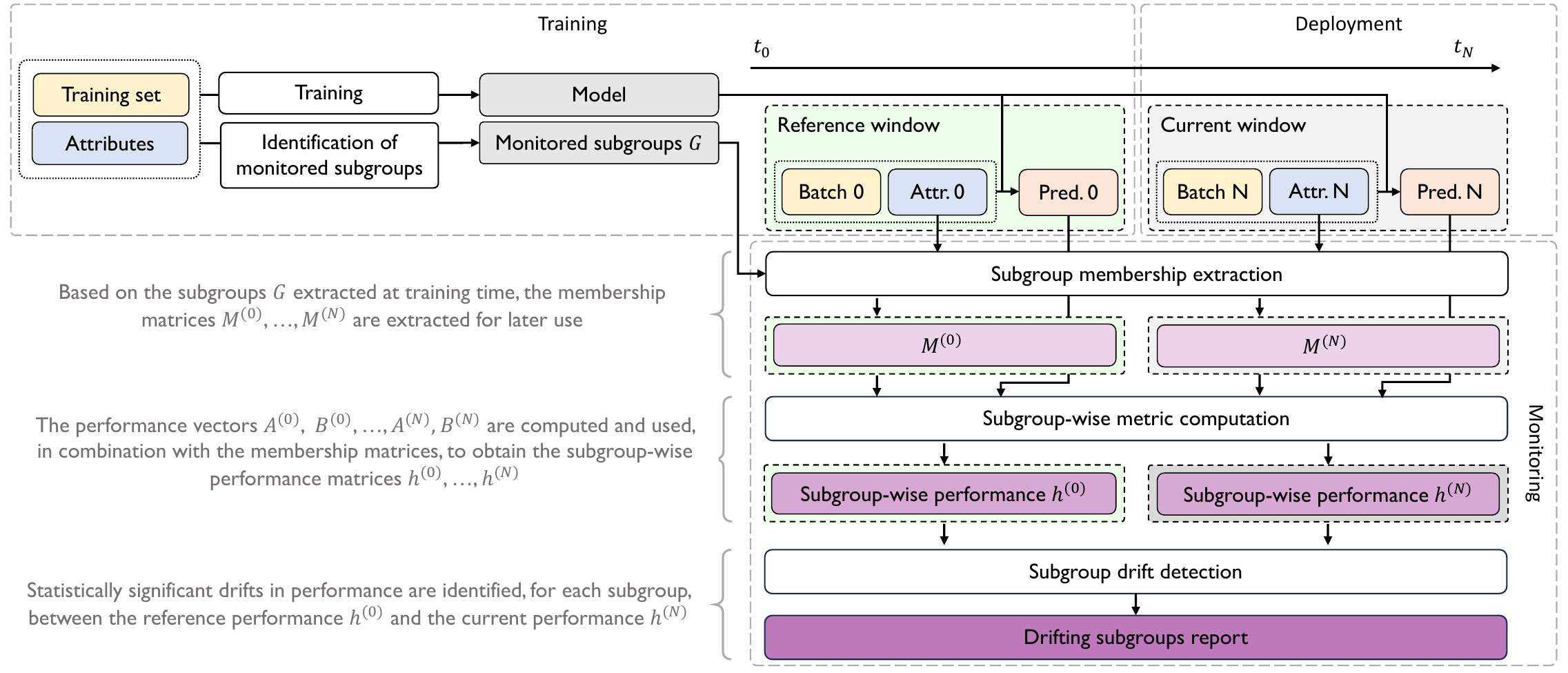}
    \caption{Main steps of the \pipeline process. }
    \Description{Main steps of the \pipeline process. }
    \label{fig:architecture}
\end{figure*}

Figure~\ref{fig:architecture} summarizes the proposed approach and contextualizes its adoption in a classic training/deployment scenario.
The pipeline features the following sub-processes of interest.
\begin{itemize}
    \item \textbf{Training.} As part of the model training phase, we extract the subgroups of interest using itemset mining approaches \cite{sagadeeva2021sliceline, pastor2021looking}. 
    We note that the rest of the pipeline is agnostic with respect to the choice of monitored subgroups. 

    \item \textbf{Deployment.} The previously built model is applied to new, unseen data. In this phase, we identify a \textit{reference window}, which represents a reference situation with data being sampled from the nominal data distribution (often, the distribution used during training). 
    % We use this window to define the reference behavior of the model. 
    The \textit{current window} instead represents the latest observed data points. We study the difference in performance, separately for each subgroup, between the current and the reference windows. 

    \item \textbf{Monitoring.} The divergences of the subgroups through time, i.e., the changes in the current performance w.r.t. the reference situation,  are computed. The final outcome of the monitoring phase is a summary defining the extent to which drift is occurring and which subgroups it is affecting.
    
\end{itemize}

%Existing approaches to analyzing subgroup performance rely on extracting subgroups, computing their performance metrics, and assessing any disparities compared to the overall data or their counterparts. 
%In our scenario, we would have to repeat the expensive process of subgroup exploration. In contrast, we propose a one-time extraction process that allows for efficient and continuous monitoring of their performance.
In the following, we first discuss the process of identifying the subgroups to monitor (\S\ref{sec:method:drift}). 
Then, we present an efficient representation (\S\ref{sec:method:efficient}) for the matrix-based computation of subgroup-wise metrics (\S\ref{sec:method:subgroup-metric-computation}). 
Finally, we define how we compute the drift in performance and its statistical significance at the subgroup level (\S\ref{sec:method:driftdetection}).

\subsection{Identification of Monitored Subgroups}
\label{sec:method:drift}

We focus on monitoring subgroups that are well-represented in the training set and interpretable.
Specifically, given a dataset  $\batch$, consisting for instance of the training data, we monitor the subgroups corresponding to the itemsets $S$ such that $\sup_{\batch}(S) \geq s$ for a user-specified minimum support $s$. 
Defining subgroups in terms of itemsets ensures their interpretability. 
Focusing on subgroups that are well-represented in the training set ensures that any performance variation over the subgroups has practical relevance. 
For a counter-example, the subgroup \{\emph{gender=male, height={100-120in}}\} may not contain any instance during training and evaluation, so monitoring its performance is pointless. 
We denote by $\mathcal{G} = \set{S : sup_\batch(S) \ge s}$ the set of \textit{frequent} subgroups in the training set $\batch$. 
Such frequent subgroups can be extracted via frequent pattern mining techniques such as the Apriori~\cite{agrawal1994fast} or FP-Growth~\cite{han2000mining} algorithms.

\subsection{Subgroup membership extraction} %assignment / determination / representation?
\label{sec:method:efficient}

To monitor model performance across subgroups, the naive approach would be to compute the performance for each subgroup $S$ separately by considering the instances $\batch(S)$ that satisfy $S$ and computing the performance metric $h$ over $\batch(S)$. 
This can be very inefficient: the number of monitored subgroups can be very large due to the combinatorial way in which itemsets (and thus subgroups) are defined. 
To make subgroup performance monitoring practical, we present an approach that leverages efficient sparse matrix multiplications to compute the performance across all subgroups simultaneously. 

Since the monitored subgroups $\mathcal{G}$ are chosen at training time (i.e., on the training set), they can be computed once and stored for later use. 
Specifically, we store the subgroups as an itemset membership matrix.
We define $G'$ as a $|\mathcal{G}| \times |\mathcal{I}|$ sparse matrix representation of $\mathcal{G}$. 
Each element $G'_{ij}$ is 1 if the item $\alpha_j \in \mathcal{I}$ is contained in subgroup $S_i$, and 0 otherwise:
\begin{equation}
    % G'_{ij} = \mathds{1}(I_i \sat \alpha_j)
    G'_{ij} = \mathds{1}(\alpha_j \in S_i) \eqpun . 
\end{equation}
Further, we let $G$ be a normalized version of $G'$, such that each column sums to 1; each element $G_{ij}$ is computed via: 
\begin{equation}
G_{ij} = \frac{G'_{ij}}{\sum_k G'_{ik}} \eqpun . 
\end{equation}
We will leverage this normalization in the efficient computation of the membership matrix for data instances, as described below.
    
Next, we keep track of the items present in each instance $x$ of a collection $\batch = x_1, \ldots, x_N$ via a \emph{point matrix} $P^{(\batch)} \in \{0, 1\}^{N \times |\mathcal{I}|}$, defined via: 
\begin{equation}
P^{(\batch)}_{ij} = \mathds{1}(\alpha_j \in x_i) \eqpun . 
\end{equation}
The point matrix $P^{(\batch)}$ is sparse, allowing for an efficient representation.
For simplicity, we omit the $(\batch)$ superscript when clear from context. 
Based on the groups matrix $G$ and the points matrix $P^{(\batch)}$, we define a membership matrix $M^{(\batch)} \in \{ 0, 1 \}^{N \times |\mathcal{G}|}$, which tracks the membership of instances into subgroups. 
The entry $M^{(\batch)}_{ij}$ indicates whether $x_i$ satisfies subgroup $S_j$:
\begin{equation}
\label{eq:membership-matrix}
M^{(\batch)}_{ij} = \mathds{1}(S_j \subseteq x_i)
\end{equation}
The matrix $M^{(\batch)}$ can be efficiently computed from $P^{(\batch)}$ and $G^\intercal$ via: 
\begin{equation} \label{eq-m}
M = \left\lfloor P^{(\batch)} G^\intercal \right\rfloor
\end{equation}
where the floor function $\lfloor \cdot \rfloor$ is applied element-wise. 
In (\ref{eq-m}), the normalization of $G$ is used along with the floor function to implement the membership check of input instances into subgroups. 
Similarly to $G$ and $P^{(\batch)}$, the matrix $M^{(\batch)}$ can be efficiently computed and stored as a sparse matrix. 
We will use the membership matrix $M^{(\batch)}$ to efficiently compute performance across subgroups.

\subsection{Subgroup-wise metric computation}
\label{sec:method:subgroup-metric-computation}

Recall that the performance function $h$ we monitor is expressed via the ratio (\ref{eq-h}) between positive outcomes and positive-plus-negative outcomes.
Letting $\alpha(\batch) = \sum_{x \in \batch} \alpha(x)$ and $\beta(\batch) = \sum_{x \in \batch} \beta(x)$, we can rewrite (\ref{eq-h}) as 
\begin{equation} \label{eq-hh}
h(\batch) = \frac{\alpha(\batch)}{\alpha(\batch) + \beta(\batch)} \eqpun . 
\end{equation}
Similarly, we express the performance over a subgroup $S$ in $\batch$ as: 
\begin{equation} \label{eq-hs}
h(\batch(S)) = \frac{\alpha(\batch(S))}{\alpha(\batch(S)) + \beta(\batch(S))} \eqpun , 
\end{equation}
where $\alpha(\batch(S))$ and $\beta(\batch(S))$ are the sums, limited to the instances that belong to the subgroup, or $\alpha(\batch) = \sum_{x \in \batch(S)} \alpha(x)$ and $\beta(\batch) = \sum_{x \in \batch(S)} \beta(x)$.
We can compute these quantities efficiently via the membership matrix $M^{(\batch)}$. 
To this end, let 
\begin{align*}
A^{(\batch)} & = (\alpha(x_1), \alpha(x_2), \dots, \alpha(x_N)) \\
B^{(\batch)} & = (\beta(x_1), \beta(x_2), \dots, \beta(x_N) 
\end{align*}
Further, define $A^{(\batch(S_j))}$, $B^{(\batch(S_j))}$ as the length-$N$ vectors with components:
%$
$$
A^{(\batch(S_j))}_i = A^{(\batch)}_i M^{(\batch)}_{ij} \qquad
B^{(\batch(S_j))}_i  = B^{(\batch)}_i M^{(\batch)}_{ij}
$$
Then, 
\begin{align*}
    \alpha(\batch(S_j)) & = 
    \textstyle \sum_{i=1}^N A^{(\batch(S_j))}_i = \sum_{i=1}^N A^{(\batch)}_i M^{(\batch)}_{ij} \\[1ex]
    \beta(\batch(S_j))  & = 
    \textstyle \sum_{i=1}^N B^{(\batch(S_j))}_i = \sum_{i=1}^N B^{(\batch)}_i M^{(\batch)}_{ij}
\end{align*}
allowing the computation of $h(\batch(S_j))$ via (\ref{eq-hs}). 
All these computations can be efficiently performed via sparse-matrix computation, yielding an efficient technique for monitoring performance over large numbers of subgroups defined over itemsets.

\subsection{Drift detection}
\label{sec:method:driftdetection}

To identify drift, we consider a reference window of data $\batch_R$, and a current window of data $\batch_C$. 
For each subgroup $S$, we can compute the drift 
\begin{equation} \label{eq-drift}
    \Delta h_S (R, C) = h(\batch_R(S)) - h(\batch_C(S)) 
\end{equation}
via the methods presented above. 
The remaining question is: are such drifts statistically significant? 
If a subgroup $S$ has small representation in $R$, or especially in $C$, where its support may not be above the chosen support threshold, $\Delta h_S (R, C)$ may be affected by statistical fluctuations, and the measured drift not indicative of a change in model performance. 

To estimate the statistical significance of drift, following \cite{pastor2021looking}, we note that we can consider the outcomes of $\alpha$ and $\beta$ as the outcomes of a Bernoulli process, where $\alpha=1$ indicates a positive outcome (or ``head'' in a coin toss), and $\beta=1$ a negative outcome (or ``tail''). 
If we take $\alpha=1$ and $\beta=1$ to be the observations, the distribution of the true performance $h(\batch(S))$, in a Bayesian sense, is distributed according to the Beta distribution with parameters $\alpha(\batch(S)) + 1$ and $\beta(\batch(S)) + 1$, corresponding to $\alpha(\batch(S))$ observed positive outcomes and $\beta(\batch(S))$ negative ones. 
The mean of such a Beta distribution is 
\[
  \mu(h, S \mid \batch) = \frac{\alpha(\batch(S)) + 1}{\alpha(\batch(S)) + \beta(\batch(S)) + 2} \eqpun , 
\]
and the variance is 
\[
  \nu(h, S \mid \batch) = \frac{(\alpha(\batch(S)) + 1) (\beta(\batch(S)) + 1)}{
    (\alpha(\batch(S)) + \beta(\batch(S)) + 2)^2 (\alpha(\batch(S)) + \beta(\batch(S)) + 3) 
  } \eqpun . 
\]
Thus, we can use Welch's t-test to compute the statistical significance $t_{h, S} (R, C)$ of the drift $\Delta h_S (R, C)$ given by (\ref{eq-drift}), as: 
\begin{equation} \label{eq-t-test}
  t_{h, S} (R, C) = \frac{|\mu(h, S \mid \batch_R) - \mu(h, S \mid \batch_C)|}{
    \sqrt{\nu(h, S \mid \batch_R) + \nu(h, S \mid \batch_C)}
  } \eqpun . 
\end{equation}
This result allows us to define a ranking among all subgroups based on the statistical significance of their divergence. 
% This represents the first result of the drifting subgroups report, the main outcome of \pipeline\!. 
Additionally, we can introduce a significance lower bound $\tau_t$, detecting drift on a subgroup $S$ only if $t_{h,S} > \tau_t$. 

Finally, we propose extracting a global detection outcome (i.e., whether, overall, the data is drifting). This binary outcome is helpful in defining whether any action should be taken. We adopt a strict detection policy whereby a dataset is said to be drifting if at least one of the subgroups contained within is drifting. The global detection function $G_t(R, C)$, producing a binary decision regarding whether dataset $C$ has drifted w.r.t. $R$ can be defined as:
\begin{equation}
\label{eq:test}
    G_t(R, C) = \begin{cases}
			 \textsc{t} & \text{if}\;\exists\; S\;s.t.\;t_{h,S} > \tau_t \\
          \textsc{f} & \text{otherwise}
		 \end{cases}
\end{equation}

% We can similarly introduce $G_\Delta$, based on a threshold $\tau_\Delta$ on $\Delta h_S$, which detects a global drift based on the change in performance.

% this can be either done by introducing a minimum significance level or by ranking all subgroups based on their  $t$-statistic, since a larget $t$-statistic implies a more significant divergence detected. We can additionally introduce a thresholding on the most divergent subgroup (as defined by the corresponding $t$) to produce a binary outcome (i.e. whether drift is occurring or not). Both these aspects characterize the \textit{drifting subgroups report}, which represents the actionable result that is the main outcome of \pipeline\!. 

\new{
% \subsection{Global detection guarantees}
We note that the \textit{global} subgroup (i.e., the subgroup containing the entire population) is one of the monitored subgrups in $\mathcal{G}$, so that the test (\ref{eq:test}) is also performed at the global level. Thus, \pipeline generalizes the behavior of global-only drift detection techniques such as DDM~\cite{gama2004learning} and HDDM~\cite{frias2014online}.
}

\section{Experimental evaluation}
\label{sec:exp}

% \subsection{Experimental setting}
\label{sec:expsetting}
We evaluate \pipeline from both the qualitative and quantitative perspectives. 
For the qualitative evaluation, we discuss the insights provided by our method on performance drift. 
The quantitative evaluation assesses \pipeline capability of (1)~detecting the occurrence of a drift event, and (2) correctly identifying the subgroups for which a drift has actually occurred. 
\new{We investigate the ability to detect both \textit{local} (i.e., subgroup-based) and \textit{global} drifts (i.e., uniformly affecting the data) of our approach.}
The source code of \pipeline\!, the code necessary to reproduce our experiments, and further details on the experimental setting are available at \githubrepo.

\subsection{Experimental setting}

\paragraph{Local drift datasets}

\begin{table}[ht]
\caption{Main features of Adult and CelebA ($s$ = 0.01).}
\label{tab:datasets}
\resizebox{\linewidth}{!}{
\begin{tabular}{rll}
\toprule
 & \textbf{Adult} & \textbf{CelebA} \\
\midrule
\textbf{\# Samples} & 48,842 & 205,599 \\
\textbf{\# Metadata} & 14 & 39 \\
%Positive class \% & 16.05\% & 51.25\% \\
%Target & Income $\ge$ 50K? & Attractive? \\
\begin{tabular}[l]{@{}c@{}}\textbf{\# Subgroups}\end{tabular} & 192,099 & 26,597 \\
\textbf{Dataset type} & Tabular & Images \& annotations \\
\textbf{Features type }& Categorical \& continuous %(binned)
& Binary \\
\bottomrule
\end{tabular}
}
\end{table}

We tested \pipeline on two commonly adopted datasets, Adult \cite{misc_adult_2} and CelebA \cite{liu2015faceattributes}, which we adapted by injecting drift (as explained below). Table \ref{tab:datasets} summarizes the main features of the two datasets. 
For Adult, we used the income as target. 
For CelebA, we identified one of the 40 binary variables as the target, and, specifically, the attribute ``Attractive''.
This attribute has been chosen as it represents a non-trivial and subjective task, that may possibly be subject to drifts over time. 
We note that we do not endorse training models for this target as something %generally 
useful or desirable beyond our experiments here. 

% Adult is a tabular dataset (with both categorical and continuous attributes) comprised of approximately 50,000 samples. CelebA instead provides approximately 200,000 images of celebrities annotated with information in the form of boolean attributes (e.g., young/old, male/female, attractive/not attractive), for a total of 40 descriptors.

% The Adult dataset already provides a binary classification problem, i.e., predicting whether the income is above or below the \$50,000/year threshold. For CelebA, we can, instead, build a binary classifier to predict, from a given image, either of the 40 target variables available. Once the target has been identified, we adopt the remaining 39 attributes as interpretable descriptors of the input image: these descriptors will be leveraged to extract the subgroups of interest. For this specific study, the binary attribute ``attractive'' has been adopted as a target, since it is the best-balanced attribute across the dataset (51.25\% of the pictures are labeled as attractive), and it represents a non-trivial and subjective task, that is possibly subject to drifts over time. However, we acknowledge the subjectivity and complexity of the topic, and do not endorse, in general, the training of models addressing such tasks.

Both datasets have been split into a 50/50 train/test split. The test split has been further split into 30 equally-sized batches. In this way, we can simulate the application of the final model to a sequence of batches through time (as is generally the case for deployed models). 

To obtain statistically representative results, we conducted multiple experiments. 
For each experiment, the data is shuffled, and the occurring drift varies (as detailed below). Each of these executions is considered as a self-contained experiment.

\paragraph{Local drift injection}
\label{par:injection}
We introduced a controlled drift that can then be detected and quantified.
To this end, we randomly selected a subgroup for each experiment. We refer to this subgroup as the \textit{target subgroup}.
Noise that affects a subset of the target subgroup is then introduced. In this way, a ground truth regarding the points affected by drift can be established. 

Noise is introduced so as to represent two main kinds of drifts. % of interest.
\begin{itemize}
    \item \textit{Concept drift.} The label for the injected points is changed (i.e., flipped in a binary classification problem). In this way, we simulate a change in behavior in the outcome of specific population subgroups, as is the case when concept drift occurs. We inject this drift in the Adult dataset. The degree of the noise can be modulated by varying the fraction of points that are affected by the label flip. 
    \item \textit{Domain drift.} Noise is applied directly to the input (e.g., Gaussian blur for images). In this way, the distribution of the input space changes, as is generally the case with domain shift. We adopt this approach for the CelebA dataset. The entity of the noise can be modulated by varying the intensity of the introduced blur. 
\end{itemize}

%For consistency throughout the experiments, 
We distribute the injection of noise in the 30 batches as follows.
\begin{itemize}
    \item \textit{Normal batches (batches 1-10).} No noise is injected, thus representing the original, expected behavior.
    \item \textit{Transition batches (batches 11-20).} A transitory phase where noise is increasingly injected in the batches. These batches represent the transitory where the drift starts occurring.
    \item \textit{Drift batches (batches 21-30).} The noise saturates to a fixed value. It is the situation in which the drift already occurred. 
\end{itemize}

 % A subset of these batches can be used as the reference window and represents the original behavior
 % We mainly focus on placing the current window in this situation. For completeness, Subsection \ref{ssec:transitory} explores \pipeline\!'s behavior during the transitory period.    

For the presented results, we consider a reference window and a current window, both containing 5 batches, taken from the normal and drift batch groups, respectively. We empirically verified that the reported results are stable for  larger and smaller batch sizes.
Appendices \ref{app:sensitivity} and \ref{app:transitory} explore the sensitivity of the proposed approach to the window size parameter.
Finally, we note that, although injected noise alters the points of a single subgroup, this effect may also propagate to other subgroups because each point typically belongs to multiple (overlapping) subgroups.

% To introduce variance in the results obtained, we conduct the same injection experiment multiple times, targeting different subgroups each time. In this way, the behavior of \pipeline can be studied as a function of the type of subgroup affected by the drift. 

\new{
\paragraph{Global drift datasets}
We additionally studied the behavior of \pipeline when the drift occurs at a global level, i.e., no specific subgroup is drifting.
This has been the focus of existing drift detection algorithms.
As such, various commonly used datasets exist in the literature to study global drifts in streaming data.
We included the following synthetic datasets: \textit{Agrawal} \cite{agrawal1993database}, with demographic data and a binary outcome on whether or not a loan is approved, \textit{SEA} \cite{street2001streaming}, with a classification target that only depends on a subset of the input features, \textit{LED} \cite{breiman2017classification}, with noisy data from a 7-segments display, and \textit{HP} (Hyperplane) \cite{hulten2001mining}, where the problem is to predict the class of a rotation hyperplane.
For these datasets, several concepts already exist between inputs and outputs. We adopt a concept drift that gradually shifts from one concept to the other by means of a sigmoid function, a commonly adopted choice \cite{montiel2021river}. We use 5,000 training points from the nominal concept and then produce 50 batches of 200 points each. We add noise to the labels of 10\% of the points to prevent the problems from being trivial. 
}

\paragraph{Experiment structure}
We identified three main experiments of interest: (1) the detection of a local (subgroup) drift event, (2) the identification of the most drifting subgroups, i.e. the subgroups for which the drifting event is occurring more extensively, and (3) the detection of a global drift event.
For all experiments, we monitor the accuracy of the model and detect a drift based on this metric (i.e., $h(\cdot)$). The drift in performance, as defined in (\ref{eq-drift}), is referred to as $\Delta_{acc}$, with its respective $t$-statistic. 
We defined two types of drift detection experiments:
\begin{itemize}
    \item \textit{Positive experiments}: each experiment represents a sequence of batches for which drift has been injected, as detailed above. For local drifts, the target subgroup injected with noise varies in each experiment.
    \item \textit{Negative experiments}: each experiment represents a sequence of batches for which no drift has been injected. 
\end{itemize}
\new{
For a given experiment, we considered the outcome to be positive (i.e., a drift was detected) if the algorithm detected a drift in at least one of the batches in the sequence, and negative otherwise. 
We evaluated the algorithms in terms of their accuracy and $F_1$ scores; additionally, we report their false positive rate (FPR) and false negative rate (FNR).\footnote{We introduced the same number of positive and negative experiments, thus producing a balanced problem.}}

% \vspace{0.2cm}\noindent % ?
We framed the identification of the subgroups that drift most as a ranking problem.
\pipeline lets us rank the subgroups according to the detected drift. 
We take the true fraction of points that we modified via noise as the \emph{relevance} $rel_k$ of the $k$-th ranked subgroup. 
We then measure the correlation between rank position and relevance, via Discounted Cumulative Gain (nDCG) \cite{wang2013theoretical}, a metric that is commonly adopted to evaluate rankings, as well as Pearson and Spearman correlations. 
% Based on the relevance of each subgroup and its ranking, the nDCG is defined as:
% \begin{equation}
%     nDCG@K = \frac{1}{c} \sum_{k=1}^K \frac{rel_k}{log(k + 1)} \eqpun , 
% \end{equation}
% %
% where $c$ is a normalizing constant representing the score obtained by the best possible ranking. It introduces an upper bound of 1 for the metric.
%This metric evaluates the overall quality of the top-$K$ results (each relevance is discounted at lower ranks).
We also compute the $nDCG@K$ to evaluate the quality of top-$K$ results.
Specifically, we evaluated nDCG@10, nDCG@100 (i.e., for the top 10 and 100 results), and nDCG for the full ranking.
Additionally, we measured the Pearson correlation between the relevance of the subgroups and the respective $t$ value, as well as the Spearman's correlation on rankings. 
% (i.e., the Pearson correlation computed on the rankings).

% We define the relevance of each subgroup $rel_j$ to be proportional to the extent to which each subgroup has been altered 

% - specify 50 exp for each support bucket

We studied the behavior of \pipeline for affected subpopulations of varying size.
The experimental results are presented as a function of the support of the target subgroup.
We produced up to 50 positive experiments for each considered support.~\footnote{If less than 50 subgroups exist for a given support, all subgroups are considered, producing less than 50 experiments.}
The average across all experiments is provided when reporting aggregated results over all supports (e.g., in Tables \ref{tab:detection-results} and \ref{tab:subgroups-ranking}).

\paragraph{Classification models}
For the Adult dataset, we adopted XGBoost \cite{chen2016xgboost} with 100 estimators as it has been shown to perform well in terms of accuracy \cite{misc_adult_2}. For CelebA, we used a pre-trained ResNet50 model with a binary classification head added to address the introduced binary problem. \new{For the synthetic datasets, we used decision trees, given the simple nature of these benchmarks.}

\paragraph{Baselines}
We compared \pipeline with seven global-level drift detection techniques. 
\new{We considered techniques that monitor model performance over time, specifically DDM~\cite{gama2004learning}, HDDM~\cite{frias2014online}, and Page-Hinkley~\cite{pagetest1954, mouss2004test}, statistical tests, specifically the Chi-squared test $\chi^2$~\cite{Pearson1900} and the Fisher’s Exact Test (FET)~\cite{fisherexact1922} and windowing-base techniques, specifically ADWIN~\cite{bifet2007learning} and Kolmogorov-Smirnov Windowing (KSWIN)~\cite{raab2020reactive}, as benchmarks, as these are among the best known and most widely used techniques.}
% During our evaluation, we vary the parameters of each method. 
% For the evaluation, we identified the best set of hyperparameters for each method and report those results. 
For DDM, we fine-tuned the minimum number of samples required before detecting a change $W$. 
For HDDM, we varied the confidence level for the drift $\epsilon$, as defined in~\cite{frias2014online}.
\new{For Page-Hinkley~\cite{pagetest1954, mouss2004test}, we varied the minimum number of samples $\lambda$ required before detecting a change. 
For $\chi^2$ and FET, we varied the p-value for the significance test.
%when comparing the distribution of the error for the reference batches with the
}
For ADWIN, we varied the significance threshold $\delta$, which determines the difference between the average performance statistics of the two sub-windows before a drift is identified as such. 
\new{For KSWIN, we varied the size of the sliding window $W$. }
%Appendix~\ref{sec:appendix:expdetails}.

%
% Given an experiment, we say that the method detected a drift if it flagged at least once during its operation after the first batch (when we start injecting bias).

\subsection{Experimental results}
\label{sec:experimentalresults}

\paragraph{Qualitative analysis}

\begin{table}
\setlength{\tabcolsep}{3pt}
\caption{Example of subgroup drift detection. Top-4 subgroups by $\mathbf{t}$ statistic and 17th subgroup (target subgroup of the experiment). CelebA dataset.}
\label{tab:topsubgroups_celeba}
\begin{tabular}{l|cc}
\toprule
\multicolumn{1}{c|}{\textbf{Subgroup}} & $\mathbf{\Delta_{acc}}$ & $\mathbf{t}$ \\ \midrule
\begin{tabular}[c]{@{}l@{}}\colorA{Big\_Lips}, \colorB{Wearing\_Lipstick}, \colorC{Wavy\_Hair}, \colorD{Young}, \\ Heavy\_Makeup, Oval\_Face, Arched\_Eyebrows \end{tabular} & -0.91 & 43.8 \\
\begin{tabular}[c]{@{}l@{}}\colorA{Big\_Lips}, \colorB{Wearing\_Lipstick},  \colorC{Wavy\_Hair}, \colorD{Young}, \\  Heavy\_Makeup, Oval\_Face \end{tabular} & -0.88 & 43.2 \\
\begin{tabular}[c]{@{}l@{}}\colorA{Big\_Lips}, \colorB{Wearing\_Lipstick},  \colorC{Wavy\_Hair}, \colorD{Young} \\ Heavy\_Makeup, Oval\_Face, No\_Beard \end{tabular} & -0.88 & 43.1 \\
\begin{tabular}[c]{@{}l@{}}\colorA{Big\_Lips}, \colorB{Wearing\_Lipstick}, \colorC{Wavy\_Hair}, \colorD{Young}, \\  Heavy\_Makeup\end{tabular} & -0.69 & 42.9 \\ \hline
\colorA{Big\_Lips}, \colorB{Wearing\_Lipstick}, \colorC{Wavy\_Hair}, \colorD{Young} & -0.62 & 39.8 \\ \bottomrule
\end{tabular}
\end{table}

To explore the insights provided by \pipelinev, we performed an experiment on the CelebA dataset. We injected noise in the subgroup \{Big\_Lips, Wearing\_Lipstick, Wavy\_Hair, Young\}. 
This subgroup has a support of 0.075 in the test set.
%Figure~\ref{fig:example-running} shows the accuracy computed on the overall test set and on the target subgroup over time. 
%While the overall accuracy changes only slightly, the target subgroup accuracy drops significantly.
%
%Over time, while the overall accuracy changes only slightly, the target subgroup accuracy drops significantly, with a trend close to the reported in Figure~\ref{fig:example-running}.
\pipeline detects this subgroup as experiencing drift, with a divergence of -0.62\% and a $t$-value of 39.8.
This subgroup is among the top 20 with the highest t-statistic (specifically, the $17^{th}$). Many of the subgroups with higher $t$-value are indeed subsets of the considered subgroup. This effect can be observed in Table~\ref{tab:topsubgroups_celeba} that lists the four subgroups with the highest $t$-value, while the bottom line reports the target subgroup.   
Hence, \pipeline is able to identify the data subgroups affected by drift. We include further qualitative analyses in the appendix.

%\pipeline highlights the data subgroups affected by drift. Practitioners can inspect the subgroups and derive a summary of the drifting behavior. Further, the interpretable definition of subgroups facilitates the inspection of the different factors associated with drift.

%\subsubsection{Overall drift detection}
\paragraph{Subgroup drift detection}
%Based on $G_t$ and $G_\Delta$, \pipeline can be used to produce a binary outcome on whether a drift has occurred in any given experiment. 
We compare the performance of \pipeline against DDM, HDDM, and ADWIN in terms of $F_1$ score when varying the support of the target subgroup. We adopt \pipeline using $\tau_t = 5$. We investigate the impact of the $\tau_t$ threshold in the last part of the section. %More detailed considerations on these parameters can be found in Appendix \ref{app:sensitivity}. 
% The computational overhead to compute the subgroup-wise metrics for a given window is approximately 5s and 10s for the CelebA and Adult datasets, respectively.
\rev{In terms of execution time, we compare \pipeline against other drift detection techniques. Under the same conditions (i.e., applying the other detection techniques on each subgroup separately), \pipeline is at least 100 times faster than then next fastest technique. More details on execution time are reported in Appendix \ref{app:time}.}

Figure~\ref{fig:overall-detection-ours-vs-rw} reports the experimental comparison of the above methods. The methods achieve near-perfect performance in identifying drifts when the target subgroup is sufficiently large (e.g., 10\% or more), because a large fraction of the entire dataset is being affected by the drift. However, \pipeline outperforms the other techniques for drift affecting smaller subgroups.
This improvement in drift detection is the result of the finer granularity of the subgroup-level analysis.
Figure \ref{fig:example-running} illustrates this: when observing global level performance, local drifting effects may be lost, as in the case with the global competitors.

%This happens thanks to the subgroup-level analysis of the change in performance: intuitively, in Figure \ref{fig:example-running} it is easier to detect the drift on the subgroup accuracy, as done by \pipeline, instead of the drift on the overall accuracy -- as done by approaches such as DDM, HDDM and ADWIN.

A summary of the performance in terms of accuracy, $F_1$ score, FPR and FNR is reported in Table \ref{tab:detection-results}, across injected subgroups of different size (i.e., with different support). 
\new{We see that most algorithms (including \pipeline\!) have low FPR: in other words, the detectors rarely report a drift when none occurred. 
The situation is different for FNR. 
Global algorithms generally have large FNR, as they fail to detect subgroup-level drifts; in contrast, \pipeline achieves distinctly lower FNR thanks to its finer-grained analysis (we note that, although KSWIN and Page-Hinkley have lower FNR on CelebA, both are affected by a much larger FPR). This indicates, as expected, that the proposed approach can detect a wider variety of drifts w.r.t. established techniques. 
% In particular, \pipeline better detects those drifts that can only be detected at the local, and not global level -- as shown in Figure \ref{fig:overall-detection-ours-vs-rw}.
}

\begin{figure}
    \centering
    \includegraphics[width=\linewidth]{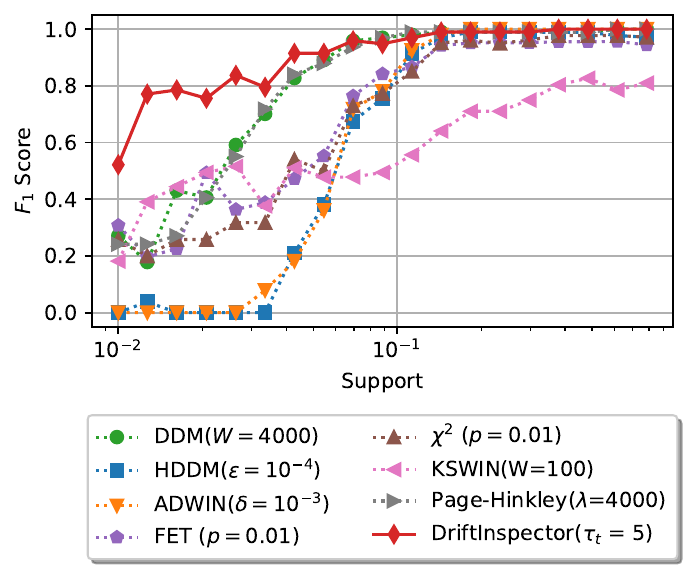}
    \caption{Drift detection performance as the target subgroup support varies; Adult.
    \pipeline achieves satisfactory results even when small portions of the population drift. 
    Each point obtained from up to 50 positive experiments (i.e., with injected drift) and the same number of negative ones.}
    \Description{Plot; drift detection performance as the target subgroup support varies; Adult dataset.}
    \label{fig:overall-detection-ours-vs-rw}
\end{figure}

\begin{table*}[ht]
\setlength{\tabcolsep}{4pt}
\caption{Accuracy (Acc.), $\mathbf{F_1}$ score, FPR, and FNR results on Adult and CelebA for the binary drift detection task. Best results for each metric. Results reported as mean $\pm$ standard deviation.}
\label{tab:detection-results}
\begin{tabular}{lrcccccccc}
\toprule
 & \multicolumn{1}{l}{} & \begin{tabular}[c]{@{}c@{}} \textbf{$\mathbf{\chi^2}$} \\ \textbf{($p = 0.01$)} \end{tabular} 
 & \begin{tabular}[c]{@{}c@{}} \textbf{ADWIN} \\ \textbf{($\delta=10^{-3}$)}\end{tabular} 
 & \begin{tabular}[c]{@{}c@{}}\textbf{DDM}  \\ \textbf{($W=4000$)}\end{tabular} 
 & \begin{tabular}[c]{@{}c@{}}\textbf{FET} \\ \textbf{($p = 0.01$)}\end{tabular} 
 & \begin{tabular}[c]{@{}c@{}}\textbf{HDDM} \\ \textbf{($\epsilon=10^{-4}$)}\end{tabular} 
 & \begin{tabular}[c]{@{}c@{}} \textbf{KSWIN} \\ \textbf{($W=100$)}\end{tabular} 
 & \begin{tabular}[c]{@{}c@{}}\textbf{Page-Hinkley} \\ \textbf{($\lambda=4000$)}\end{tabular} 
 & \begin{tabular}[c]{@{}c@{}}\textbf{\pipeline} \\  \textbf{($\tau_t = 5$)}\end{tabular} \\   \midrule
 %
 % & \multicolumn{1}{l}{} & \rotatebox[origin=c]{90}{\begin{tabular}[c]{@{}c@{}} \textbf{$\chi^2$} \\ \textbf{($p = 0.01$)} \end{tabular}} 
 % & \rotatebox[origin=c]{90}{\begin{tabular}[c]{@{}c@{}} \textbf{ADWIN} \\ \textbf{($\delta=10^{-3}$)}\end{tabular}} 
 % & \rotatebox[origin=c]{90}{\begin{tabular}[c]{@{}c@{}}\textbf{DDM}  \\ \textbf{($W=4000$)}\end{tabular}} 
 % & \rotatebox[origin=c]{90}{\begin{tabular}[c]{@{}c@{}}\textbf{FET} \\ \textbf{($p = 0.01$)}\end{tabular}} 
 % & \rotatebox[origin=c]{90}{\begin{tabular}[c]{@{}c@{}}\textbf{HDDM} \\ \textbf{($\epsilon=10^{-4}$)}\end{tabular}} 
 % & \rotatebox[origin=c]{90}{\begin{tabular}[c]{@{}c@{}} \textbf{KSWIN} \\ \textbf{(W=100)}\end{tabular}} 
 % & \rotatebox[origin=c]{90}{\begin{tabular}[c]{@{}c@{}}\textbf{Page-Hinkley} \\ \textbf{($\lambda$=4000)}\end{tabular}} 
 % & \rotatebox[origin=c]{90}{\begin{tabular}[c]{@{}c@{}}\textbf{\pipeline} \\ \textbf{($\tau_t = 5$)} \end{tabular}} \\ 
\multirow{2}{*}{\rotatebox[origin=c]{90}{\textbf{Acc.}}}  & \textbf{Adult} & 0.774 \footnotesize{$\pm$ 0.185} & 0.773 \footnotesize{$\pm$ 0.223} & 0.865 \footnotesize{$\pm$ 0.163} & 0.776 \footnotesize{$\pm$ 0.175} & 0.760 \footnotesize{$\pm$ 0.221} & 0.596 \footnotesize{$\pm$ 0.119} & 0.861 \footnotesize{$\pm$ 0.168} & \textbf{0.919 \footnotesize{$\pm$ 0.092}} \\
 & \textbf{CelebA} & 0.812 \footnotesize{$\pm$ 0.144} & 0.879 \footnotesize{$\pm$ 0.170} & 0.906 \footnotesize{$\pm$ 0.090} & 0.828 \footnotesize{$\pm$ 0.119} & 0.854 \footnotesize{$\pm$ 0.152} & 0.561 \footnotesize{$\pm$ 0.022} & 0.709 \footnotesize{$\pm$ 0.074} & \textbf{0.991 \footnotesize{$\pm$ 0.014}} \\ \midrule
\multirow{2}{*}{\rotatebox[origin=c]{90}{\textbf{$F_1$}}} 
& \textbf{Adult} & 0.670 \footnotesize{$\pm$ 0.305} & 0.580 \footnotesize{$\pm$ 0.442} & 0.797 \footnotesize{$\pm$ 0.272} & 0.689 \footnotesize{$\pm$ 0.282} & 0.571 \footnotesize{$\pm$ 0.435} & 0.577 \footnotesize{$\pm$ 0.173} & 0.789 \footnotesize{$\pm$ 0.282} & \textbf{0.902 \footnotesize{$\pm$ 0.124}} \\
 & \textbf{CelebA} & 0.786 \footnotesize{$\pm$ 0.201} & 0.801 \footnotesize{$\pm$ 0.316} & 0.892 \footnotesize{$\pm$ 0.123} & 0.811 \footnotesize{$\pm$ 0.161} & 0.804 \footnotesize{$\pm$ 0.257} & 0.691 \footnotesize{$\pm$ 0.013} & 0.775 \footnotesize{$\pm$ 0.058} & \textbf{0.991 \footnotesize{$\pm$ 0.015}} \\ \midrule
\multirow{2}{*}{\rotatebox[origin=c]{90}{\new{\textbf{FPR}}}} & \textbf{Adult} & 0.074 \footnotesize{$\pm$ 0.013} & \textbf{0.000 \footnotesize{$\pm$ 0.000}} & 0.018 \footnotesize{$\pm$ 0.006} & 0.100 \footnotesize{$\pm$ 0.005} & 0.024 \footnotesize{$\pm$ 0.010} & 0.415 \footnotesize{$\pm$ 0.035} & 0.018 \footnotesize{$\pm$ 0.006} & 0.016 \footnotesize{$\pm$ 0.008} \\
 & \textbf{CelebA} & 0.178 \footnotesize{$\pm$ 0.058} & \textbf{0.000 \footnotesize{$\pm$ 0.000}} & 0.067 \footnotesize{$\pm$ 0.029} & 0.176 \footnotesize{$\pm$ 0.046} & 0.083 \footnotesize{$\pm$ 0.028} & 0.861 \footnotesize{$\pm$ 0.041} & 0.566 \footnotesize{$\pm$ 0.142} & \textbf{0.000 \footnotesize{$\pm$ 0.000}} \\ \midrule
\multirow{2}{*}{\rotatebox[origin=c]{90}{\new{\textbf{FNR}}}} & \textbf{Adult} & 0.377 \footnotesize{$\pm$ 0.363} & 0.454 \footnotesize{$\pm$ 0.446} & 0.253 \footnotesize{$\pm$ 0.326} & 0.348 \footnotesize{$\pm$ 0.349} & 0.456 \footnotesize{$\pm$ 0.445} & 0.394 \footnotesize{$\pm$ 0.256} & 0.260 \footnotesize{$\pm$ 0.334} & \textbf{0.146 \footnotesize{$\pm$ 0.181}} \\
 & \textbf{CelebA} & 0.198 \footnotesize{$\pm$ 0.266} & 0.243 \footnotesize{$\pm$ 0.340} & 0.121 \footnotesize{$\pm$ 0.182} & 0.168 \footnotesize{$\pm$ 0.231} & 0.208 \footnotesize{$\pm$ 0.304} & 0.017 \footnotesize{$\pm$ 0.019} & \textbf{0.015 \footnotesize{$\pm$ 0.025}} & 0.018 \footnotesize{$\pm$ 0.028} \\ \bottomrule
\end{tabular}
\end{table*}

%\subsubsection{Drifting subgroups identification}
\paragraph{Drifting subgroups identification}
Based on the fraction of each subgroup that has been altered, Table \ref{tab:subgroups-ranking} quantifies the quality of the ranking produced by ordering the subgroups according to the $t$-statistic computed. We compare this performance with the one achieved with a random guess, since the previously considered competitors do not produce subgroup-level detections.

\begin{table}
\caption{Evaluation of the rankings, based on $\mathbf{t}$, against a random baseline.
%Experiments conducted on injected subgroups of varying size. 
Note that the theoretical random guess with Pearson and Spearman correlations is 0 -- the reported results are obtained from empirical evidence and are consistent with the expected value. 
Results reported as mean $\pm$ standard deviation.}
\label{tab:subgroups-ranking}
\resizebox{\linewidth}{!}{\begin{tabular}{crcc}
\toprule
 &  & \textbf{\pipeline} & \textbf{Random} \\
\midrule
\multirow[c]{2}{*}{\textbf{nDCG@10}} & \textbf{Adult} & \textbf{0.707 \footnotesize{$\pm$ 0.235}} & 0.151 \footnotesize{$\pm$ 0.167} \\
 & \textbf{CelebA} & \textbf{0.975 \footnotesize{$\pm$ 0.112}} & 0.305 \footnotesize{$\pm$ 0.235} \\
\cline{1-4}
\multirow[c]{2}{*}{\textbf{nDCG@100}} & \textbf{Adult} & \textbf{0.697 \footnotesize{$\pm$ 0.219}} & 0.156 \footnotesize{$\pm$ 0.161} \\
 & \textbf{CelebA} & \textbf{0.952 \footnotesize{$\pm$ 0.124}} & 0.308 \footnotesize{$\pm$ 0.225} \\
\cline{1-4}
\multirow[c]{2}{*}{\textbf{nDCG}} & \textbf{Adult} & \textbf{0.931 \footnotesize{$\pm$ 0.068}} & 0.825 \footnotesize{$\pm$ 0.088} \\
 & \textbf{CelebA} & \textbf{0.973 \footnotesize{$\pm$ 0.050}} & 0.884 \footnotesize{$\pm$ 0.073} \\
\cline{1-4}
\multirow[c]{2}{*}{\textbf{Pearson}} & \textbf{Adult} & \textbf{0.555 \footnotesize{$\pm$ 0.236}} & 0.000 \footnotesize{$\pm$ 0.002} \\
 & \textbf{CelebA} & \textbf{0.736 \footnotesize{$\pm$ 0.174}} & 0.000 \footnotesize{$\pm$ 0.006} \\
\cline{1-4}
\multirow[c]{2}{*}{\textbf{Spearman}} & \textbf{Adult} & \textbf{0.457 \footnotesize{$\pm$ 0.225}} & 0.000 \footnotesize{$\pm$ 0.002} \\
 & \textbf{CelebA} & \textbf{0.676 \footnotesize{$\pm$ 0.217}} & 0.000 \footnotesize{$\pm$ 0.006} \\
\bottomrule
\end{tabular}}
\end{table}

% vecchia considerazione non piu' applicable
% Interestingly, empirical evidence shows that the unnormalized difference in performance (Eq.~\ref{eq-drift}) produces a better ranking, with respect to the corresponding t-statistic (Eq.~\ref{eq-t-test}). We observe this behavior consistently across datasets and metrics.

The nDCG@10 and nDCG@100 metrics show that, even for small values of $K$, \pipeline already highlights the most relevant results. Since manual inspection of the most drifting subgroups is generally expected, this is a result of particular interest.

\begin{figure}
    \centering
    \includegraphics[width=\linewidth]{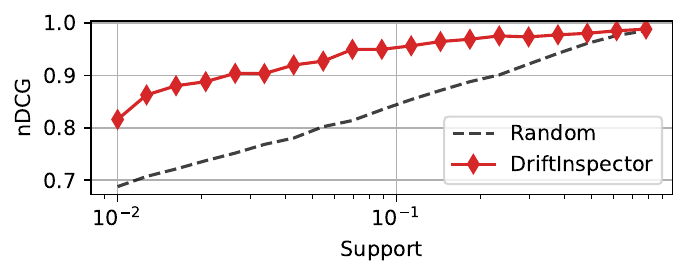}
    \caption{nDCG varying the target subgroup support; Adult.
    %dataset.
    }
    \Description{Plot; nDCG score as a function of the target subgroup support. Adult dataset. }
    \label{fig:ndcg-vs-sup}
\end{figure}

In Figure \ref{fig:ndcg-vs-sup}, we explore the quality of the ranking as a function of the size of the target subgroup. 
For drifts with support below $10\%$, the proposed approach already produces meaningful rankings, as highlighted by the difference with respect to the random baseline.
When the support of the affected subgroup increases, more instances are perturbed and more subgroups are affected by noise on average. Hence, the fraction of affected points in most subgroups will become larger. 
In this case, as can be expected, \pipeline produces near-perfect rankings. Note that, for support values close to 100\%, almost all instances are affected by noise and random selection also reaches an ``almost perfect'' result.

%We note that the random baseline also improves in performance when larger subgroups are injected with noise. 
%As the support increases, more instances are perturbed; therefore, more subgroups are affected by noise on average. Consequently, the relevance (i.e., the fraction of points affected) of most subgroups will grow toward larger values. This introduces ``high relevance'' results in high-rank positions of the randomly ranked subgroups. 

\new{
\paragraph{Global drift detection.} We additionally studied the behavior of \pipeline when the drift occurs at a global level, i.e., there is no one specific subgroup drifting. This has been the focus of existing drift detection algorithms.
As such, we compared \pipeline with  existing techniques on four benchmark datasets. Table \ref{tab:global-drift} presents the main results obtained in terms of accuracy, $F_1$ score, FPR and FNR. 
\pipeline generally obtains comparable, or better performance w.r.t. other techniques when global drift occurs. 
This same behavior can be observed on the right-hand side of Figure \ref{fig:overall-detection-ours-vs-rw}, when the injected subgroup approximately encompasses the entire dataset (i.e., a global drift occurs). 

\begin{table}[ht]
    \centering
    \caption{
    \new{Performance of various drift detection techniques on 4 synthetic datasets, with global drift applied. %Best results shown in bold for each metric.
    }}
    \label{tab:global-drift}
    \resizebox{\linewidth}{!}{
\begin{tabular}{lrcccccccc}
\toprule
 &  & \rotatebox[origin=c]{90}{$\mathbf{\chi^2}$} & \rotatebox[origin=c]{90}{\textbf{ADWIN}} & \rotatebox[origin=c]{90}{\textbf{DDM}} & \rotatebox[origin=c]{90}{\textbf{FET}} & \rotatebox[origin=c]{90}{\textbf{HDDM}} & \rotatebox[origin=c]{90}{\textbf{KSWIN}} & \rotatebox[origin=c]{90}{\textbf{Page-Hinkley}} & \rotatebox[origin=c]{90}{\textbf{\pipeline}} \\ \midrule
\multirow{4}{*}{\rotatebox[origin=c]{90}{\new{\textbf{Accuracy}}}} & \textbf{Agrawal} & 0.986 & 0.986 & 0.943 & 0.986 & 0.986 & 0.886 & \textbf{1.000} & \textbf{1.000} \\
 & \textbf{LED} & 0.486 & 0.500 & 0.471 & 0.457 & 0.486 & \textbf{0.514} & 0.471 & 0.500 \\
 & \textbf{SEA} & 0.700 & 0.843 & 0.529 & 0.543 & 0.543 & 0.443 & 0.571 & \textbf{1.000} \\
 & \textbf{HP} & 0.757 & 0.914 & 0.871 & 0.771 & 0.914 & 0.614 & \textbf{0.929} & 0.886 \\ \midrule
\multirow{4}{*}{\rotatebox[origin=c]{90}{\new{\textbf{$F_1$ score}}}} & \textbf{Agrawal} & 0.986 & 0.986 & 0.943 & 0.986 & 0.986 & 0.885 & \textbf{1.000} & \textbf{1.000} \\
 & \textbf{LED} & 0.327 & 0.357 & 0.320 & 0.385 & 0.470 & 0.508 & 0.394 & \textbf{0.545} \\
 & \textbf{SEA} & 0.689 & 0.840 & 0.431 & 0.440 & 0.510 & 0.327 & 0.490 & \textbf{1.000} \\
 & \textbf{HP} & 0.755 & 0.914 & 0.871 & 0.771 & 0.914 & 0.590 & \textbf{0.928} & 0.897 \\ \midrule
\multirow{4}{*}{\rotatebox[origin=c]{90}{\new{\textbf{FPR}}}} & \textbf{Agrawal} & 0.029 & 0.029 & 0.114 & 0.029 & 0.029 & 0.029 & \textbf{0.000} & \textbf{0.000} \\
 & \textbf{LED} & \textbf{0.029} & \textbf{0.029} & 0.057 & 0.886 & 0.686 & 0.371 & 0.171 & 0.447 \\
 & \textbf{SEA} & 0.114 & 0.029 & 0.057 & 0.029 & 0.714 & 0.143 & 0.029 & \textbf{0.000} \\
 & \textbf{HP} & 0.143 & \textbf{0.000} & 0.057 & 0.171 & 0.114 & 0.629 & 0.143 & 0.103 \\ \midrule
\multirow{4}{*}{\rotatebox[origin=c]{90}{\new{\textbf{FNR}}}} & \textbf{Agrawal} & \textbf{0.000} & \textbf{0.000} & \textbf{0.000} & \textbf{0.000} & \textbf{0.000} & 0.200 & \textbf{0.000} & \textbf{0.000} \\
 & \textbf{LED} & 1.000 & 0.971 & 1.000 & \textbf{0.200} & 0.343 & 0.600 & 0.886 & 0.562 \\
 & \textbf{SEA} & 0.486 & 0.286 & 0.886 & 0.886 & 0.200 & 0.971 & 0.829 & \textbf{0.000} \\
 & \textbf{HP} & 0.343 & 0.171 & 0.200 & 0.286 & 0.057 & 0.143 & \textbf{0.000} & 0.129 \\
\bottomrule
\end{tabular}}
\end{table}

}

\begin{figure}
    \centering
    \includegraphics[width=\linewidth]{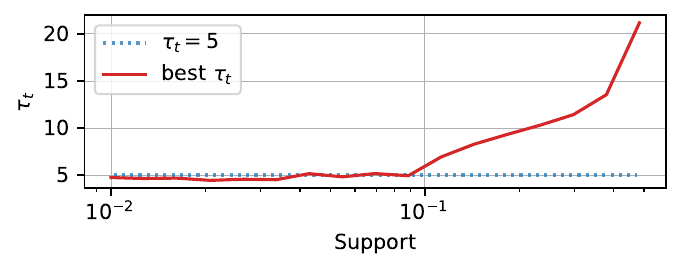}
    \caption{Evolution of the best threshold value  for $\mathbf{\tau_t}$ on Adult, when varying the target subgroup support. The threshold value $\mathbf{\tau_t = 5}$ used in the experiments is shown in blue.}
    \Description{Plot; evolution of the best threshold value for $\mathbf{\tau_t}$ on Adult, when varying the target subgroup support.}
    \label{fig:tau-t-vs-sup}
\end{figure}

\paragraph{Support sensitivity analysis.}
%\paragraph{Sensitivity analysis of the $\tau_t$ threshold for drift detection}
%The identification of a threshold value for $\tau$ (either $\tau_t$ or $\tau_\Delta$) is crucial for the binary drift detection problem. 
We analyze the impact of the $\tau_t$ threshold on the performance of the drift detection task.
More specifically, we study how the optimal $\tau_t$ changes with the support of the injected subgroup.
We define the optimal threshold as the one maximizing Youden's J statistic \cite{youden1950index} ($J = TPR + TNR - 1$, where TPR and TNR are the true positive and negative rates). 
Figure~\ref{fig:tau-t-vs-sup} shows the value of the optimal threshold when varying subgroup support.
The optimal $\tau_t$ value, corresponding to $\tau_t \approx 5$, is stable for supports < 10\%. This range, corresponding to smaller subgroups, is the focus of this work. Still, Figure~\ref{fig:overall-detection-ours-vs-rw} shows that $\tau_t = 5$ allows achieving good results also for larger supports, even though it is not the optimal threshold. Larger supports correspond to prominent drifts, for which there is tolerance in the setting of the threshold.

% Similar considerations on $\tau_\Delta$ are presented in Appendix~\ref{app:sensitivity}.

%Table \ref{fig:subgroupdrift:ranking} with NDCG (and Pearson) for ours and random baseline.
%Figure \label{fig:subgroupdrift-support-ndcg-corr}

%\subsubsection{Sensitivity analysis}
%In appendix!

%%%%%%%%%%%%%%%%%%%%%%%%%%%%%%%%%%
% Overall
% Adult, CelebA
% Methods: Ours vs DDM/Adwin
% Metrics: Accuracy
% Curve accuracy vs support for Ours, DDM, Awin
% Let's see varying the subgroup
% We observe the existing methods fail to identify drift for smaller subgroups. 
% Similar conclusions for CelebA
% Example for an injected drift: Overall drift not detectable
%%%%%%%%%%%%%%%%%%%%%%%%%%%%%%%%%%%%
% Qualitative
%%%%%%%%%%%%%%%%%%%%%%%%%%%%%%%%%%%%
%%%%
% Subgroup
% Adult, CelebA
% Methods: Ours vs Subgroup+Random vs 
% Clusters sul training? -- evitare for the moment 
% Metrics: NGDG + Other metric of ranking
%%%%%%%%%%%%%%%%%%%%%%%%%%%%%%%%%%%%
%%% Sensitivity analysis
% Parameter and ROC
% Curva per un sottogruppo del noise e del t/divergence
%%%%%%%%%%%%%%%%%%%%%%%%%%%%%%%%%%%%

% Experiments to run
% CelebA overall - result preparation
% Run drift

\section{Conclusions}
\label{sec:conclusion}
We propose a novel approach to explore drift phenomena at the finer granularity of subgroups. We automatically identify relevant subgroups during model training and efficiently monitor their drift during model deployment. Our approach allows detecting finer-grained, but highly drifted, data subsets that are not detected by methods that observe drift as a global phenomenon. \pipeline also allows us to explore the collection of (interpretable) drifting subgroups, thereby serving as a tool for drift understanding and enabling actively working on model improvement.
In future work, we plan to explore subgroup-level detection of data distribution shifts when labeled data is unavailable.

%\begin{acks}
%\end{acks}

\bibliographystyle{ACM-Reference-Format}
\bibliography{main}

\appendix
\newpage

\section{Further qualitative analysis}

We conduct further qualitative analyses of \pipeline results to illustrate its effectiveness as a tool for drift inspection and model understanding.
We again consider the example discussed in Section \ref{sec:experimentalresults}.
In this example, we inject noise for the subgroup \{Big\_Lips, Wearing\_Lipstick, Wavy\_Hair, Young\}.
\rev{We analyze (i) a summary of the subgroups, (ii) the contribution to the drift of specific subgroups (local contribution), and
(iii) the contribution of each metadata to the drift (global contribution). }

\paragraph{Subgroup summary.}
\rev{
We can generate a summarized view of subgroup divergence to analyze subgroups more easily. We follow an approach close to the redundancy pruning approach of~\cite{pastor2021looking}. The rationale is the following: given an itemset $I$ and an itemset $I \cup \alpha$, where $\alpha$ represents an additional item, we keep the more general $I$ if the difference in their $t$-value falls below a predefined threshold. Essentially, $I$ already represents the divergence of $I \cup \alpha$, with the additional term $\alpha$ having only minimal impact. Different from~\cite{pastor2021looking}, we prune based on the $t$-value as a measure of the strength of the difference between groups rather than the divergence.
Table~\ref{tab:topsubgroups_celeba_red} shows the subgroups with the highest $t$-values after applying a redundancy pruning threshold of 5. These pruned subgroups contain fewer terms. The target subgroup is the second by $t$-value. The other subgroups include metadata from the target subgroup and others associated with them, such as `Heavy\_Makeup' to `Wearing\_Lipstick'.}

\begin{table}[ht]
\setlength{\tabcolsep}{3pt}
\caption{\rev{Summary of the top-5 subgroups by $t$ statistic via redundancy pruning of 5 in $t$-value. CelebA dataset.}}
\label{tab:topsubgroups_celeba_red}
\begin{tabular}{l|cc}
\toprule
\multicolumn{1}{c|}{\textbf{Subgroup}} & $\mathbf{\Delta_{acc}}$ & $\mathbf{t}$ \\ \midrule
\colorA{Big\_Lips}, \colorC{Wavy\_Hair}, \colorD{Young}, Heavy\_Makeup & -0.68 & 42.33 \\
\colorA{Big\_Lips}, \colorB{Wearing\_Lipstick}, \colorC{Wavy\_Hair}, \colorD{Young} & -0.62 & 39.76 \\
\colorA{Big\_Lips}, \colorC{Wavy\_Hair}, Oval\_Face, Arched\_Eyebrows  & -0.88 & 36.93 \\
\colorA{Big\_Lips}, \colorC{Wavy\_Hair}, Heavy\_Makeup  & -0.60 & 36.32 \\
\colorA{Big\_Lips}, \colorB{Wearing\_Lipstick}, \colorC{Wavy\_Hair} & -0.54 & 34.65 \\ \bottomrule
\end{tabular}
\end{table}

\

\begin{figure}[ht]
    \centering
    \includegraphics[width=0.9\linewidth]{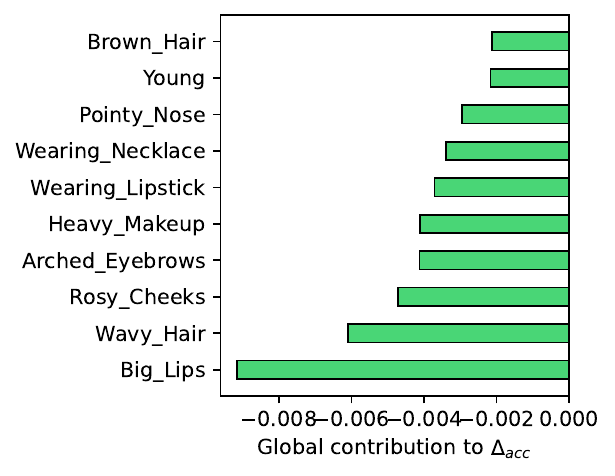}
    \caption{Global contribution to the drift in accuracy. Top 10 terms with the highest contributions to drift. CelebA dataset. }
    \Description{Bar plot; global contribution to the drift in accuracy}
    \label{fig:global_shapley_value_celebA}
\end{figure}

\paragraph{\rev{Global contribution to the drift}}
To investigate the drift phenomenon, % further,
we can explore the contribution of each item (i.e., metadata) to the drift. Using the Global Shapley Value~\cite{pastor2021looking}, we can estimate the average contribution of each item to the drift, considering its effect on all explored subgroups.
Figure~\ref{fig:global_shapley_value_celebA} shows the top 10 terms with the highest contribution. The higher the magnitude of the value, the more the item impacts the performance reduction. 
`Wavy\_Hair' and `Big\_Lips' are the most influential metadata contributing to drift. `Wearing\_Lipstick' also stands out as a predominant term. `Young' instead exhibits a lower contribution, likely due to its high frequency in the dataset (80\%).
An interesting observation is the prevalence of multiple terms with high negative contribution among images labeled as female in the dataset (e.g., `Rosy\_Cheeks',  `Arched\_Eyebrows'). 
This impact of gender-associated metadata (for this dataset) aligns with observing that almost all images in the target subgroup are labeled as women (99.78\%).

Practitioners can inspect the subgroups and derive a summary of the drifting behavior. Further, the interpretable definition of subgroups facilitates the inspection of the different factors associated with drift.

\begin{figure}[ht]
    \centering
    \includegraphics[width=\linewidth]{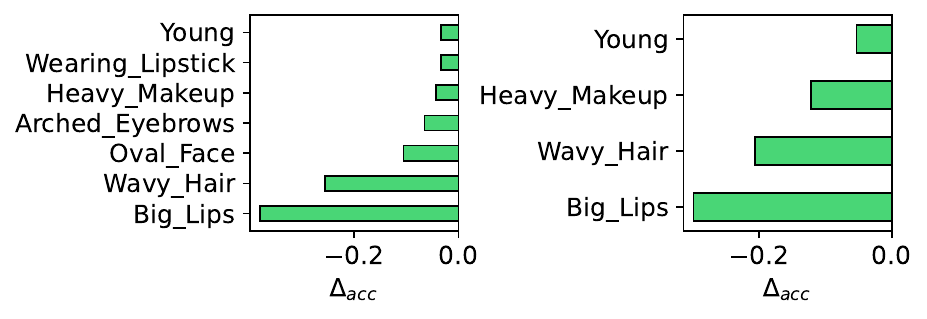}
    \caption{Contribution to the drift in accuracy for the subgroup with the highest $\mathbf{t}$-value and for the target dataset. CelebA dataset.}
    \Description{Bar plots; contribution to the drift in accuracy for the subgroup with the highest $\mathbf{t}$-value and for the target dataset. CelebA dataset }
    \label{fig:shapley_value_celebA}
\end{figure}

\paragraph{Local contribution to the drift.}
The subgroup with the highest $t$-value in the change in performance is \{Big\_Lips, Wearing\_Lipstick, Oval\_ Face, Wavy\_Hair, Arched\_Eyebrows, Young, Heavy\_Makeup\} (see Table \ref{tab:topsubgroups_celeba}).
The subgroup includes all metadata from the target one plus additional terms. As a result, the instances of this subgroup are all included in the target one and are affected by the noise injection.

We are interested in analyzing how much each term contributes to the drift. For this purpose, we can use the Shapley value, as adopted in \cite{pastor2021looking}.
The Shapley value is a notion from game theory that estimates the contribution of each team player to the total score. In our context, we consider the divergence in the performance of a subgroup as the total score, the subgroup as the team, and each item (i.e., each metadata) as the players.
Figure \ref{fig:shapley_value_celebA} (left) shows the Shapley values for the drifting subgroups with the highest $t$-value. The terms `Big\_Lips' and `Wavy\_Hair' are the terms that mostly contribute to the drift in performance for the subgroup. We then have the terms `Oval\_Face', `Arched\_Eyebrows', `Heavy\_Makeup'. 
As we analyzed, % in Section \ref{tab:topsubgroups_celeba}, 
these terms are prevalent among images labeled as `woman'. Consequently, their relevance is justified by the fact that almost all images in target subgroups are labeled as `woman' as well (98.8\%).

Figure \ref{fig:shapley_value_celebA} (right) shows the contributions for the target subgroups. The most important terms are again `Big\_Lips' and `Wavy\_Hair'. The term `Young' has the lowest contribution in absolute terms. Given that most of the images of the entire dataset (80\%) are labeled as `Young', it is not a distinctive characteristic of the subgroups affected by noise. This justifies the higher contributions of the other metadata.

\begin{figure}
    \centering
    \includegraphics[width=\linewidth]{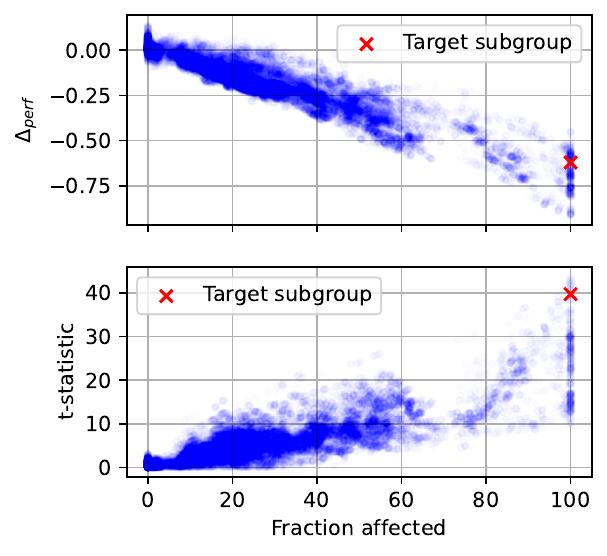}
    \caption{
    Correlation between the fraction of points that have been altered for each subgroup, and the detection of drift, in terms of $\mathbf{\Delta_{acc}}$ and $\mathbf{t}$-statistic.  }
    \Description{Two scatter plots; correlation between the fraction of points that have been altered for each subgroup, and the detection of drift, in terms of $\mathbf{\Delta_{acc}}$ and $\mathbf{t}$-statistic. }
    \label{fig:correlation-alterated-div-t}
\end{figure}

By injecting noise for images of the target subgroup, we are consequently modifying instances of other overlapping subgroups as well.
We study the relationship between the fraction of affected instances in each subgroup and its divergence and $t$-value.
Figure~\ref{fig:correlation-alterated-div-t} shows the divergence scores (top) and $t$-values (bottom) of the explored subgroups with respect to the fraction of altered instances in each subgroup.
We observe a clear trend.
The higher the fraction of altered points, the lower the divergence (and higher in absolute terms), indicating a drop in accuracy for the subgroups. Similarly, higher fraction correspond to higher $t$-values.
In the figures, we highlight the target subgroup. As also discussed in Section~\ref{sec:experimentalresults}, there are subgroups that experience even higher drops. These subgroups are subsets of the target one (e.g., Big\_Lips, Wearing\_Lipstick,  Wavy\_Hair, Young,  Heavy\_Makeup, Oval\_Face, see Table~\ref{tab:topsubgroups_celeba}).

\section{Sensitivity analysis}
\label{app:sensitivity}
\begin{figure}[ht]
    \centering
    \includegraphics[width=0.7\linewidth]{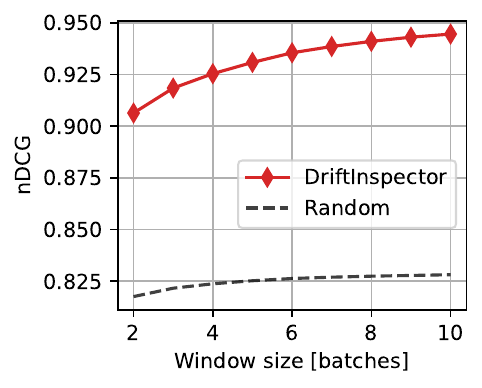}
    \caption{nDCG as the window size changes, on the Adult dataset.}
    \Description{Plot; nDCG as the window size changes, on the Adult dataset.}
    \label{fig:winsize-adult}
\end{figure}

\begin{figure}
    \centering
    \includegraphics[width=0.7\linewidth]{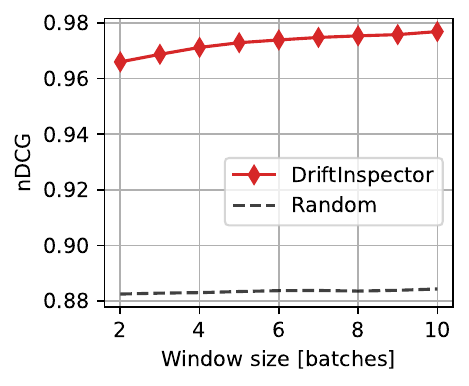}
    \caption{nDCG as the window size changes. CelebA dataset.}
    \Description{Plot; nDCG as the window size changes. CelebA dataset.}
    \label{fig:winsize-celeba}
\end{figure}

% results: 
% window size , on nDCG. 
\paragraph{Window size definition}
For the main results of the experimental section, we adopted a window covering 5 batches. This parameter introduces a trade-off between the significance of the measured drift (larger windows will encompass more samples, thus providing a more meaningful estimate of the drift), and the delay introduced. We study the entity of the delay in Appendix \ref{app:transitory}. We evaluate the ranking performance, in terms of nDCG, as the window size increases in Figures \ref{fig:winsize-adult} and \ref{fig:winsize-celeba}. In both cases, we observe that even smaller window sizes already produce high quality results (as compared against the random baseline). A slight improvement in performance can be observed as the window size increases: since this improvement is less meaningful after 5 batches, we choose this as the window size for all experiments. We note, however, that this parameter is affected by factors such as size of the batches and velocity of the drift. As such, case by case considerations should be made when defining this parameter.


\section{Complexity analysis}
\rev{
This section presents a complexity analysis quantifying the number of operations required for each iteration (batch).

The initial cost for the extraction of the subgroups of interest (at training time) depends on the algorithm of choice. We consider this cost as separate from the computational cost required by \pipeline for the extraction of the subgroup-wise metrics.

We assume that a total of $|\mathcal{G}|$ subgroups are being monitored, and that a batch of size $N$ is being considered. We additionally refer to $u$ as the average number of items (metadata) used to describe each point, %(for tabular metadata, this is the number of attributes/columns), 
whereas $l$ represents the average length of each subgroup, in terms of items. In the general case, we expect $u$, $l << |\mathcal{I}|$. 

Based on this notation, we report below the time and space complexities of each step of \pipeline\!.

The computation of the sparse matrix $G$ is $O(|\mathcal{G}| \cdot l)$ in both space (to store the matrix in sparse format) and time. Similarly, the computation of the sparse matrix $P$ is $O(N \cdot u)$ in space and time.

The computation of the membership matrix $M$ is the result of the multiplication between sparse matrices. For each of the dot products between rows of $P$ and columns of $G^\intercal$, only non-zero pairs of values need to be computed. To compute the total number of products that need to be computed, we need to estimate the number of non-zero pairs of values. 
We make the simplifying assumption of independence between $P$ and $G^\intercal$ (we note, however, that this assumption is not always applicable, depending on the technique used for the extraction of the subgroups). For a given row of $P$ and column of $G^\intercal$, the expected number of non-zero pairs is of 
$\frac{u}{|\mathcal{I}|} \cdot \frac{l}{|\mathcal{I}|} \cdot |\mathcal{I}|$ non-zero pairs. Thus, the time required for the sparse multiplication if $O(N \cdot |\mathcal{G}| \cdot  \frac{u \cdot l}{ |\mathcal{I}|})$. Based on the sparsity of the problem (as defined by $u$, $l$), this is generally much lower than the cost required for the non-sparse multiplication $O(N \cdot  |\mathcal{G}| \cdot |\mathcal{I}|)$. 
It is more difficult to compute the space complexity, since it requires considerations on the relationship between points and subgroups. The lower bound can be established (assuming using a frequent pattern mining with a support threshold \minsup) as $N \cdot |\mathcal{G}| \cdot  \minsup$. The upper bound is $N \cdot |\mathcal{G}|$, where all points belong to all subgroups (i.e., $\minsup = 1$). We argue that this is, however, an extreme case where subgroups do not convey any useful information. 

The $\alpha$, $\beta$ vectors require $O(|\mathcal{G}|)$ in space and between $N \cdot |\mathcal{G}| \cdot \minsup$ and $N \cdot |\mathcal{G}|$ in time for their computation. 

Finally, the computation of the t-statistics for each subgroup is $O(|\mathcal{G}|)$ in time and space. 

The overall time cost in the worst case will be $N \cdot |\mathcal{G}|$, whereas in the best case it will be $N \cdot |\mathcal{G}| \cdot  (\frac{u \cdot l} {|\mathcal{I}|} \cdot s)$ . The overall space complexity will be between $N \cdot |\mathcal{G}|$ (worst case scenario) and $N \cdot |\mathcal{G}| \cdot \minsup$ (best case). 
}

\section{Transitory}
\label{app:transitory}
The main results presented focus on the identification and characterization of the drift batches, as defined in the experimental section. However, understanding the behavior of \pipeline during the transitory period (transition batches) is also of interest. 

\begin{figure}[ht]
    \centering
    \includegraphics[width=\linewidth]{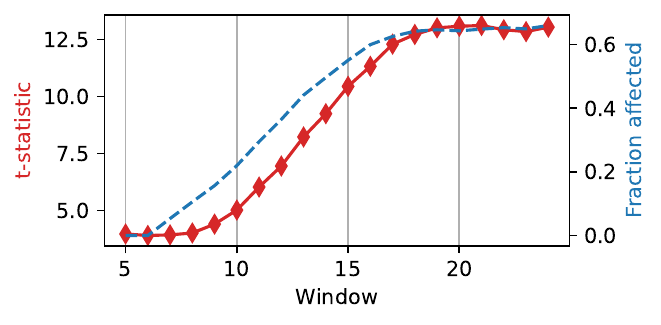}
    \caption{Drift indicators ($\mathbf{\Delta_{acc}}$ and $\mathbf{t}$) as the current window slides over the 30 test batches. The window size used is of 5 batches. $\mathbf{-\Delta_{acc}}$ is reported for consistency with the rest of the plot. Y values for $\mathbf{-\Delta_{acc}}$ and $\mathbf{t}$ normalized to the same range for visualization purposes.}
    \Description{Plot; drift indicators ($\mathbf{\Delta_{acc}}$ and $\mathbf{t}$) as the current window slides over the 30 test batches}
    \label{fig:transitory}
\end{figure}

In Figure \ref{fig:transitory} we qualitatively observe how the detection of divergence varies as the \textit{current} window moves forward in time.

The reference window includes batches 1 to 5.
The x axis represents the batch at which each ``current'' window starts: for instance, Windows start = 6 represents the window that starts immediately after the reference window.
The blue, dashed line represents the fraction of altered points for the target subgroup in the current window. 
The other lines represent the drift of the most diverging subgroup, as detected by the proposed approach.
The figure highlights the inertia of \pipeline that leads to a delay between when the drift begins and when it is detected. We note that one of the parameters that can affect this inertia is the size of the current/reference reference windows.

\section{Experimental setting details}
\label{sec:appendix:expdetails}

This section outlines the performance-based approaches at the overall level against which we compare our model. 
\new{Specifically, we adopt DDM~\cite{gama2004learning}, HDDM~\cite{frias2014online}, Page-Hinkley~\cite{mouss2004test, pagetest1954}, two algorithm based on sliding windows, ADWIN~\cite{bifet2007learning} and KSWIN~\cite{raab2020reactive} and the statistical tests $\chi^2$ and FET. 
For the first 5 methods, we use the implementation available in the \textit{scikit-multiflow}~\cite{skmultiflow} library; for $\chi^2$ and FET, we use the \textit{alibi-detect}~\cite{alibi-detect}
library.}
DDM~\cite{gama2004learning} detects a change if there is a significant increase in the error rate, specifically if the sum of the error rate and standard deviation at time $t$ is greater or equal to the sum of the minimum error rate and three times the minimum standard deviation. DDM also includes a parameter, the minimum number of samples, to prevent false detections. This parameter determines the minimum number of samples that need to be analyzed before a change can be detected. We consider the values [500, 1000, 2000, 4000, 8000]. 
%The method achieved the best results with a value of 4000.
%

HDDM~\cite{frias2014online} is a modification of DDM that uses Hoeffding's inequality~\cite{hoeffding1994probability} to detect changes in performance.
We tested its variant HDDM\_A, which uses the average as estimator. %designed for sudden drifts, and HDDMW for gradual drifts.
We vary the confidence for the drift level, considering the values [0.0001, 0.0005, 0.001, 0.002, 0.004, 0.05, 0.1]. 
%We report the best result with value 0.0001.
%

\new{Page-Hinkley~\cite{mouss2004test, pagetest1954} monitors the cumulative sum of the deviations of the observed values from their mean based on the Page-Hinkley test\cite{pagetest1954}.
At each time step, the approach computed the deviation of the observed value at time $t$ from the mean considering the value up to time $t$ and added to the cumulative sum.
The test maintains a minimum value of this cumulative sum to identify the point of maximum deviation.
If the difference between the current cumulative sum and the minimum is greater than a given threshold, it indicates a significant change in the mean, suggesting a drift.
The implementation of Page-Hinkley requires the minimum number of instances to be observed before the algorithm evaluates if there is a change in the data distribution. }

\new{FET and $\chi^2$ drift detectors are two non-parametric drift detectors. They apply the 
 applies Fisher’s Exact Test (FET)~\cite{fisherexact1922} and the 
$\chi^2$ Test~\cite{Pearson1900}, respectively.
We compare the error distribution for the reference batches without nominal behavior with new sequence batches. We varied the p-value for the significance test, considering values.} % [0.001, 0.01, 0.25, 0.05].

ADWIN~\cite{bifet2007learning} is an adaptive sliding window algorithm that automatically keeps a variable-length window and divides it into two sub-windows to determine if a change has happened. ADWIN compares the average performance statistic over these two windows and detects a drift if their difference surpasses a pre-defined significance threshold. 
We vary the significance threshold of the difference for flagging a drift. We consider the values [0.001, 0.002, 0.004, 0.05, 0.1].
%and obtain the best results for 0.001.

\new{KSWIN~\cite{raab2020reactive} (Kolmogorov-Smirnov Windowing) is a sliding-window algorithm that maintains two sliding windows over the data stream and compares their distribution with the Kolmogorov-Smirnov test. 
A window is a reference window representing the past data, while the other is the current one representing the recent data.
The approach computes the Kolmogorov-Smirnov statistic between the data distributions in the two windows to test whether the two samples come from the same distribution. 
If the KS statistic exceeds a given threshold, it indicates a significant difference between the distributions of the two windows and, hence, a potential drift.
In the experiment, we monitor the error and we use the default value for the threshold of the test statistic (value 0.005).
To define the two windows, 
KSWIN defines the two windows based on two parameters: the size of the sliding window $W$ and the size of the statistic window $r$. 
In a sliding window of size $W$, the last $r$ points represent the most recent window, while from the first $W-r$ sample, it uniformly draws $r$ samples to define the reference window.
In the experiments, we vary the window of size $W$ considering the values 100 to 800 with a step of 100, while we set $r$ at the default value of 30.} % For the real-world dataset Adult and CelebA, we obtain the best result for $W$ equal to 700.

\begin{figure}[ht]
    \centering
    \includegraphics[width=0.7\linewidth]{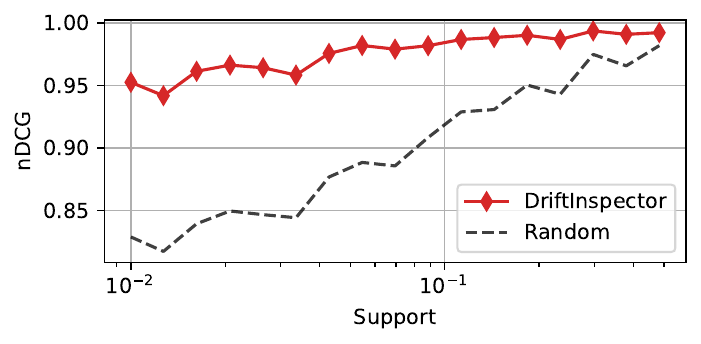}
    \caption{nDCG as the support of the target subgroup varies, on the CelebA dataset.}
    \Description{}
    \label{fig:ndcg-vs-sup-celeba}
\end{figure}

\begin{figure}[ht]
    \centering
    \includegraphics[width=0.7\linewidth]{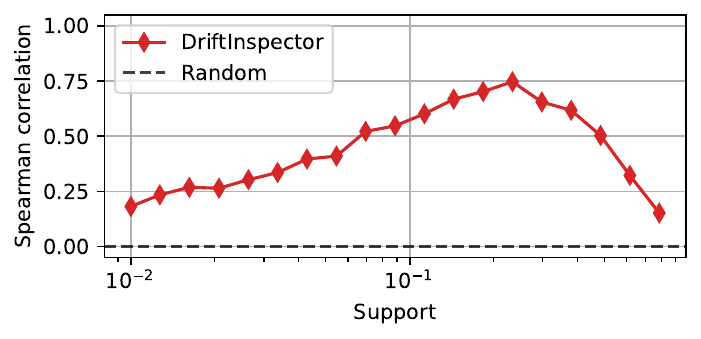}
    \caption{Spearman's correlation as the support of the target subgroup varies, on the Adult dataset.}
    \Description{}
    \label{fig:spearman-vs-sup-adult}
\end{figure}

\begin{figure}[ht]
    \centering
    \includegraphics[width=0.7\linewidth]{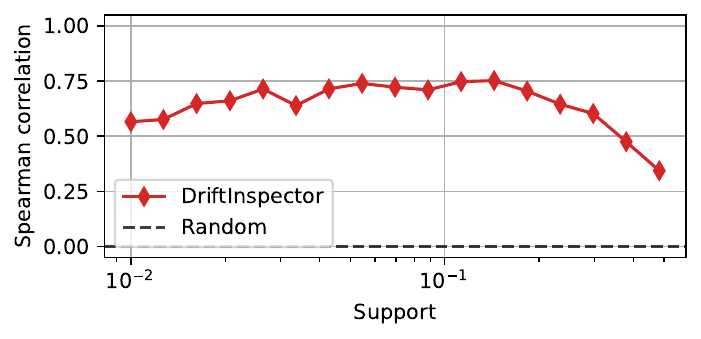}
    \caption{Spearman's correlation as the support of the target subgroup varies, on the CelebA dataset.}
    \Description{}
    \label{fig:spearman-vs-sup-celeba}
\end{figure}

\begin{figure}[ht]
    \centering
    \includegraphics[width=0.7\linewidth]{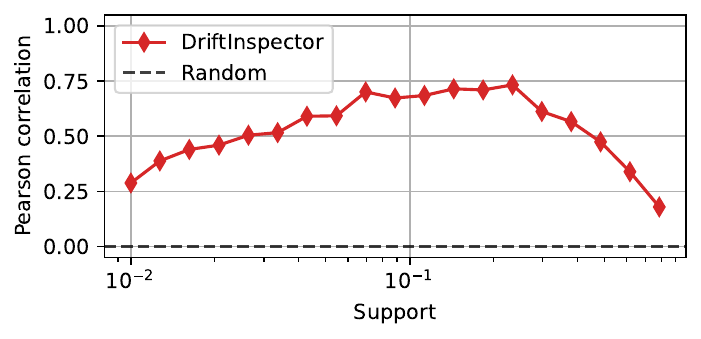}
    \caption{Pearson correlation as the support of the target subgroup varies, on the Adult dataset.}
    \Description{}
    \label{fig:pearson-vs-sup-adult}
\end{figure}

\begin{figure}[ht]
    \centering
    \includegraphics[width=0.7\linewidth]{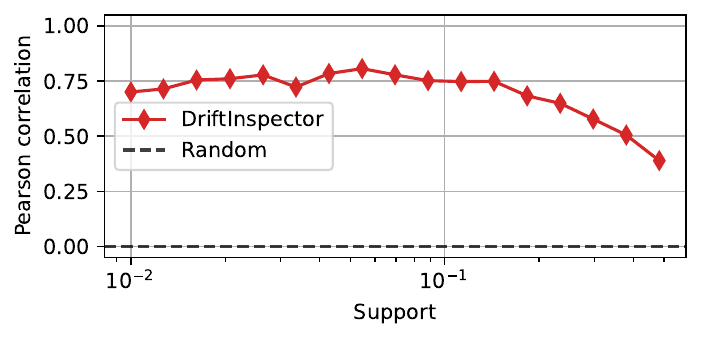}
    \caption{Pearson  correlation as the support of the target subgroup varies, on the CelebA dataset.}
    \Description{}
    \label{fig:pearson-vs-sup-celeba}
\end{figure}

\section{Further results on drifting subgroup identification}
\label{sec:appendix:furtherranking}
We additionally present the ranking results on Adult in terms of Pearson (Figure \ref{fig:pearson-vs-sup-adult}) and Spearman's correlation (Figure \ref{fig:spearman-vs-sup-adult}) between the estimated $\Delta_{acc}$ (or $t$) and the fraction of altered points in each subgroup.

We observed a decrease in Pearson and Spearman's correlation for larger supports. This behavior occurs, as already discussed in the experimental section, because when most subgroups drift, the fraction of altered points for most groups rises toward 100\%. This saturation results in a less meaningful correlation with the extracted drift indicator, thus producing a lower correlation.

For completeness, we also studied the nDCG (Figure \ref{fig:ndcg-vs-sup-celeba}), Pearson (Figure \ref{fig:pearson-vs-sup-celeba}) and Spearman's correlation (Figure \ref{fig:spearman-vs-sup-celeba}) as a function of the size of the target subgroup, for the CelebA dataset. The same considerations that have been made for Adult also apply for CelebA.

\section{Time comparison}
\label{app:time}
\rev{In this Section we compare the execution time of various detection algorithms, when computed separately for each subgroup, against that of \pipeline\!. 

For comparability of the results, we take each detector and apply it to each subgroup separately (i.e., a separate detector is created for each subgroup). This approach would constitute a baseline method to be used as a competitor for \pipeline\!. However, the execution times of this naive technique makes the performance comparison unfeasible. 
As such, we only report the test times, as measured on a limited number of batches. 

The results are reported in Table \ref{tab:time}. Thanks to the intrinsic support of subgroups, \pipeline is approximately 100 times faster than the next fastest algorithms ($\chi^2$ and Fisher's Exact Test), and as much as 2,000 faster than KSWIN. }

% \usepackage{multirow}
% \usepackage{multirow}
\begin{table}[ht]
\caption{\rev{Execution times, normalized by the sample size, for the subgroup-wise detection of drifts. Each column represents a separate value for the minimum support \minsup of the subgroups (i.e., a different number of subgroups being monitored). }}
\label{tab:time}
\resizebox{\linewidth}{!}{
\begin{tabular}{rccc}
\toprule
\multicolumn{1}{l}{\multirow{2}{*}{}} & \multicolumn{3}{c}{\textbf{Execution time (s)}} \\
\multicolumn{1}{l}{} & \minsup = 0.01 & \minsup = 0.05 & \minsup = 0.1 \\ \midrule
$\chi^2$     & 1.2867 $\pm$ 0.0251 & 0.1249 $\pm$ 0.0026 & 0.0359 $\pm$ 0.0005 \\
ADWIN        & 2.1340 $\pm$ 0.0140 & 0.2095 $\pm$ 0.0033 & 0.0610 $\pm$ 0.0020 \\
DDM          & 1.8865 $\pm$ 0.0242 & 0.1804 $\pm$ 0.0053 & 0.0519 $\pm$ 0.0031 \\
FET          & 1.2876 $\pm$ 0.0406 & 0.1224 $\pm$ 0.0021 & 0.0350 $\pm$ 0.0003 \\
HDDM         & 5.2180 $\pm$ 0.2554 & 0.4991 $\pm$ 0.0066 & 0.1423 $\pm$ 0.0037 \\
KSWIN        & 25.1528 $\pm$ 0.9003 & 2.5874 $\pm$ 0.0184 & 0.7769 $\pm$ 0.0255 \\
Page-Hinkley & 2.3472 $\pm$ 0.0847 & 0.2204 $\pm$ 0.0025 & 0.0657 $\pm$ 0.0023 \\
\pipeline    & \textbf{0.0120 $\pm$ 0.0001} & \textbf{0.0012 $\pm$ 0.0000} & \textbf{0.0004 $\pm$ 0.0000} \\ \bottomrule
\end{tabular}}
\end{table}

\end{document}